\documentclass{article} % For LaTeX2e
\usepackage{iclr2027_conference,times}

\usepackage{amsmath,amsfonts,bm}

\def\eqref#1{equation~\ref{#1}}
\def\1{\bm{1}}

\DeclareMathAlphabet{\mathsfit}{\encodingdefault}{\sfdefault}{m}{sl}
\SetMathAlphabet{\mathsfit}{bold}{\encodingdefault}{\sfdefault}{bx}{n}

\newcommand{\E}{\mathbb{E}}

\usepackage{hyperref}
\usepackage{url}

\usepackage{times}
\usepackage{soul}
\usepackage[utf8]{inputenc}
\usepackage[small]{caption}
\usepackage{graphicx}
\usepackage{amsmath}
\usepackage{amsthm}
\usepackage{booktabs}
\usepackage{algorithm}
\usepackage{algorithmic}
\usepackage[switch]{lineno}
\usepackage{amsfonts,amssymb}
\usepackage{stfloats}
\usepackage{multirow}
\usepackage{makecell}
\usepackage{graphicx}
\usepackage{cleveref}

\usepackage{colortbl}
\definecolor{light-gray}{gray}{0.82}
\newcolumntype{g}{>{\columncolor{light-gray}}c}
\newcommand*\rot{\rotatebox{55}}
\newcommand{\graylabel}[1]{\begingroup\setlength{\fboxsep}{1pt}\colorbox{light-gray}{#1}\endgroup}

\usepackage{xcolor}
\usepackage{tcolorbox}
\usepackage[T1]{fontenc}
\usepackage{textcomp}
\usepackage{enumitem}

\title{Beyond Natural Images: Rethinking AI-\\Generated Image Detection in Documents}

\author{\textbf{Zhangjie Fu}$^{1}$ \quad
\textbf{Jiazhen Yan}$^{1}$ \quad
\textbf{Yuanwen Chen}$^{2}$ \quad
\textbf{Xinquan Yu}$^{3}$ \quad \\
\textbf{Yanzhe Li}$^{2\dagger}$ \quad
\textbf{Hui Jiang}$^{2,4}$ \quad
\textbf{Lei Gao}$^{2\S}$ \quad
\textbf{Chenfu Bao}$^{2,4\S}$ \quad
\\
$^{1}$Engineering Research Center of Digital Forensics, Ministry of Education, Nanjing University of \\ Information Science and Technology \quad
$^{2}$Baidu Inc. \quad \\
$^{3}$School of Computer Science and Engineering, MoE Key Laboratory of Information Technology, \\ Guangdong Province Key Laboratory of Information Security Technology, Sun Yat-sen University  \quad \\
$^{4}$Tsinghua University \quad \\
\texttt{fzj@nuist.edu.cn} \quad \texttt{247918horizon@gmail.com} \\
{\footnotesize $^\dagger$Project Leader. \, $^\S$Corresponding Author.}
}

\iclrfinalcopy % Uncomment for camera-ready version, but NOT for submission.
\begin{document}

\maketitle

\begin{abstract}
AI-generated image detection has attracted increasing attention, but existing evaluations mainly focus on natural images, leaving AI-generated document images largely underexplored. This omission is concerning because documents often appear in sensitive real-world scenarios, such as invoices, expense reports, certificates, and medical records. In this paper, we first construct a controlled diagnostic benchmark, \textbf{AIGDoc-Pilot}, and reveal that existing detectors suffer substantial performance degradation on AI-generated document images, with the mean AUC dropping by more than 7\%. Based on this, we further reveal two document-specific properties behind this gap: generation artifacts exhibit strong spatial inconsistency across local regions, and text density significantly affects real-synthetic separability, where text-dense regions offer stronger discriminative evidence. Motivated by these findings, we construct \textbf{AIGDoc}, a larger document-centric dataset containing diverse real-world documents and AI-generated counterparts produced by multiple advanced generation and editing models. Extensive experiments on AIGDoc demonstrate that existing detectors still struggle to reliably identify AI-generated documents, while document-based training partially narrows the gap. Together, these results offer valuable insights for developing dependable and generalizable detectors in document-centric scenarios. \textbf{The code and datasets will be made publicly available upon acceptance of the paper.}
\end{abstract}

\section{Introduction}

With the continuous advancement of AI-generated image technologies \cite{ye2023ip,wu2024infinite}, realistic images can now be produced with greater ease and efficiency, thereby enriching people’s daily lives. However, this development has also made such images increasingly difficult to distinguish from authentic ones, posing significant potential risks to society, the economy, and politics \cite{wang2024iterative,wang2025pair,wang2026map,hu2023invisible,hu2026high}. Many researchers have begun to focus on training highly generalizable detectors for AI-generated images, including methods based on artifact representations \cite{li2025pay,lin2025standing,yan2026dual} and pre-trained models \cite{zhou2025breaking,liu2025beyond}, which have been applied effectively to real-world scenarios. Despite their promising performance in real-world scenarios, existing studies predominantly focus on natural images, while risks posed by AI-generated document images remain largely overlooked.

Traditionally, document forgery has often relied on manual editing tools such as Photoshop. Many studies have explored this problem within the framework of document tampering detection \cite{luo2025toward,qu2026textshield,qu2026detect}. However, recent generative models can synthesize entire document images or modify existing ones with high visual realism, introducing forgery patterns that may not be adequately captured by conventional tampering detection methods. These cases blur the boundary between document tampering detection and AI-generated image detection, and naturally raise a critical question: \textit{Can existing detectors really cover AI-generated document images?}

\begin{figure*}[h]
    \centering
    \includegraphics[width=1\linewidth]{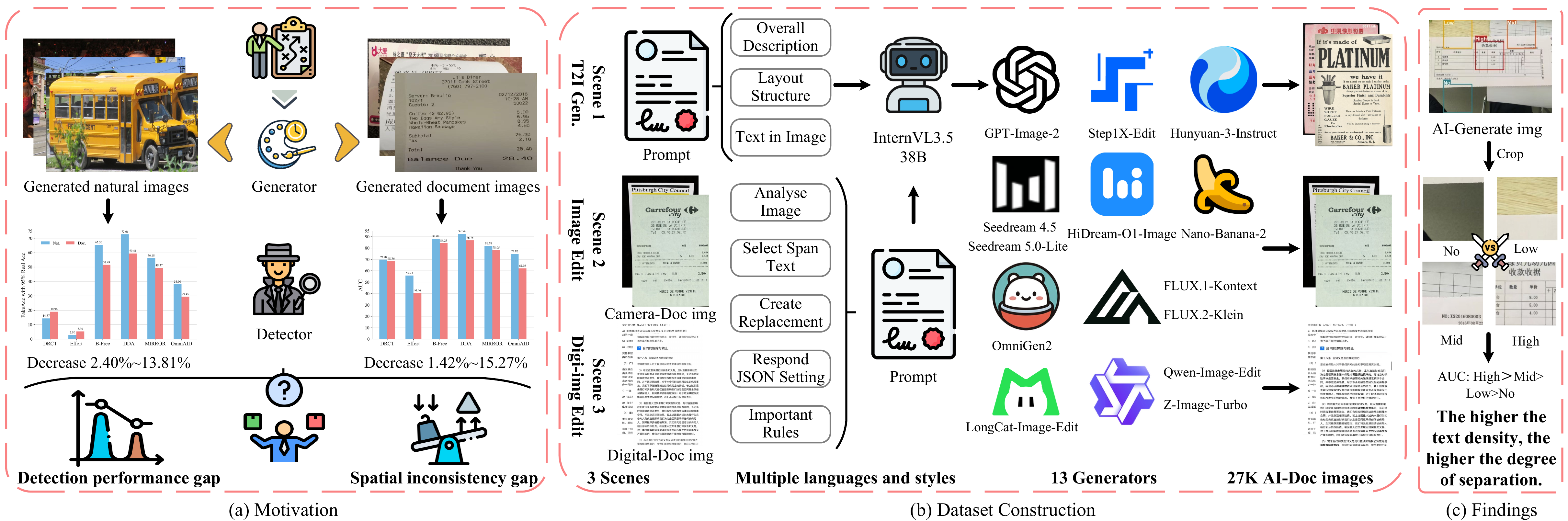}
    \caption{\textbf{(a) Diagnosis and Findings.} Using the controlled diagnostic benchmark AIGDoc-Pilot, we reveal that existing detectors suffer substantial performance degradation on AI-generated document images. Further diagnosis shows that generated document artifacts exhibit strong spatial inconsistency across local regions, while text-dense regions provide stronger discriminative evidence for real-synthetic separability. \textbf{(b) Datasets and Evaluations.} We construct AIGDoc, a large-scale benchmark for AI-generated document image detection, comprising 27K synthetic images across 17 subsets, covering 13 generation methods from three categories.}
    \label{fig:overall}
\end{figure*}

To comprehensively investigate this problem, we first construct a pilot benchmark, termed \textbf{AIGDoc-Pilot}. It contains real-world camera-captured \footnote{Existing detectors are typically trained on camera-captured images. Accordingly, digital images, rendered images, and similar data are regarded as cross-type images. In this study, we restrict the dataset to camera-captured images to ensure source consistency.} document images from multiple practical scenarios, together with their AI-generated counterparts produced by image-to-image generation under different noise levels. For comparison, we also collect an equal number of camera-captured natural images and generate corresponding synthetic images using the same pipeline. Experimental results demonstrate that existing detection methods encounter significant challenges when evaluated on AIGDoc-Pilot. Key insights are summarized as follows: 
\textbf{i) A pronounced gap between natural and document images.} Under identical generation conditions, the same detectors suffer a substantial accuracy drop on AI-generated document images compared with AI-generated natural images.
\textbf{ii) Strong spatial inconsistency of artifacts in AI-generated document images.} Compared with natural images, AI-generated document images demonstrate significantly weaker spatial coherence of generation artifacts. Detectable forgery evidence is often concentrated in limited local regions, whereas others contain only subtle artifact traces, resulting in substantial discrepancies in detection performance across distinct image patches.
\textbf{iii) Text density enhances authentic-synthetic separability.}
We observe that regions with denser textual content enable detectors to better distinguish AI-generated document images from real ones, suggesting that discriminative artifact cues in document images are closely associated with high-frequency character boundaries and layout structures.

Meanwhile, to facilitate systematic evaluation and alleviate the observed limitations of existing AI-generated image detectors on document images, we propose the \textbf{AIGDoc} dataset, which includes diverse real-world document images and AI-generated samples produced by multiple advanced generation and editing models. We show that incorporating document images during training can partially reduce the natural-document gap, underlining the need for document-centric datasets, evaluation protocols, and detection methods. Overall evaluation and datasets are shown in Figure \ref{fig:overall}.

The main contributions of this paper are as follows:
\begin{itemize}

\item To the best of our knowledge, this is the first systematic investigation of AI-generated image detection in document-centric scenarios. Through AIGDoc-Pilot, we reveal a pronounced performance gap between natural images and document images, highlighting an overlooked limitation of existing AI-generated image detectors.

\item Our findings show that generation artifacts in document images exhibit strong spatial inconsistency, and text-dense regions provide stronger discriminative artifact cues for distinguishing real and AI-generated documents. The simple text-density-guided preprocessing during training significantly boosted model performance, surpassing the state-of-the-art method by 8.9\% in mAcc.

\item We construct AIGDoc, a large-scale AI-generated document image dataset comprising 27k AI-generated images across 17 subsets, covering 13 generation methods from three categories. AIGDoc provides a practical benchmark for systematic evaluation, helping bridge the gap between AI-generated image detectors and document-centric detection scenarios.

\end{itemize}

\section{Related Work}

\subsection{AI-Generation Image Detection}
The rapid advancement of AI generative models has introduced serious risks to visual authenticity. To address these concerns, recent studies have focused on developing AI-generated image detectors with strong generalization ability. Artifact-based approaches typically exploit generation traces such as frequency-domain anomalies~\cite{luo2021generalizing,tan2024frequency,yan2026dual}, upsampling artifacts~\cite{tan2024rethinking}, local patch inconsistencies~\cite{chen2021attentive,cavia2024real}, and reconstruction residuals~\cite{wang2023dire,chen2024drct,chu2025fire}. In recent years, leveraging the transferable and robust representations of pre-trained models has become an important direction for improving detection performance, which includes feature-based methods~\cite{ojha2023towards,koutlis2024leveraging,zhang2025towards,zhou2025brought} and fine-tuning-based methods~\cite{liu2024mixture,tan2025c2p,yan2025ns,yan2025dgs,liu2026mirror}. For example, VIB-Net~\cite{zhang2025towards} applies a variational information bottleneck to suppress task-irrelevant information, NS-Net~\cite{yan2025ns} constructs a semantic NULL space to remove semantic interference from visual features, and DGS-Net~\cite{yan2025dgs} introduces distillation-guided gradient surgery to suppress harmful gradients while preserving transferable pre-trained representations. In addition, several studies improve generalization from the data perspective, such as data augmentation~\cite{li2025improving} and the construction of less biased training datasets \cite{chen2025dual,guillaro2025bias}. Despite these efforts and the promising generalization achieved on natural images, AI-generated document image detection remains largely unexplored.

\subsection{AI-Counterfeited Document Image Detection}
With the maturation of image-editing tools such as Photoshop, many researchers have devoted considerable effort to document tampering detection, in which high-frequency inconsistencies around tampering boundaries are often exploited for both detection and localization \cite{luo2025toward,song2025cross,qu2026textshield,qu2026omni,qu2026detect}, achieving promising performance. However, the rise of AI-generated content \cite{wu2025qwenimage,wu2025omnigen2,blackforestlabs2025fluxkontext} has greatly expanded editing capabilities beyond conventional manual manipulation. In particular, image editing can remove the sharp boundary artifacts and local high-frequency inconsistencies that traditional tampering detectors rely on, posing a challenge to existing document forgery detection methods.

Notably, existing AI-generated image detection research has rarely considered document images, leaving a substantial gap in the evaluation of document-centric counterfeits. To address this gap, we conduct the controlled diagnostic study on \textbf{AIGDoc-Pilot} and introduce the larger real-world benchmark \textbf{AIGDoc}. 
The most closely related work is TextFake~\cite{zhang2026textfake}, which reports detector failures on AI-generated text-rich images but mixes diverse image types and generation settings, making it difficult to determine whether the observed degradation comes from the document domain itself. In contrast, our work constructs a controlled comparison under the same generation pipeline, thereby enabling a direct measurement and analysis of the document-domain gap. We hope that our study can help bridge the gap between natural-image-oriented and document-centric forgery detection, providing a foundation for future research on reliable AI-generated document image detection.

\section{Diagnostic Study on AI-Generated Document Images}

\subsection{AIGDoc-Pilot Datasets}
\noindent\textbf{Real Image Collection.}
We observe that the training data used by existing AI-generated image detectors mainly consist of camera-captured natural images, such as MSCOCO \cite{lin2014microsoft} and ImageNet \cite{deng2009imagenet}. In contrast, document images are more diverse in acquisition forms, including camera-captured images, screenshots, and born-digital electronic documents. Directly comparing such heterogeneous document images with camera-captured natural images may introduce confounding cross-domain factors and lead to inaccurate performance evaluation. Therefore, we collect candidate document images from SRD \cite{Auad2024}, M6Doc\cite{Cheng_2023_CVPR}, and the ICDAR competition dataset \cite{yu2023icdar}. We then manually filter out images that are not captured by cameras, resulting in 545 camera-captured document images, which covers multiple document types such as books, receipts, and tickets. Meanwhile, for comparison, we randomly select the same number of natural images from Open Images V7 as the reference natural-image set. Under this constraint, both natural and document images share the same acquisition modality, where the primary difference lies in the visual content rather than the acquisition source. 

\noindent\textbf{AI-Generated Image Construction.}
To preserve the visual distribution of the original images, we construct AI-generated images using image-to-image generation. Specifically, we employ recent generative models, including HiDream-E1-Full \cite{hidreami1technicalreport}, FLUX.1-dev \cite{blackforestlabs2024flux1dev}, Qwen-Image \cite{wu2025qwenimage}, and SDv3.5-Large \cite{esser2024scaling}. Within each subset, every generated image is paired with its corresponding source real image. To reduce bias caused by a single noise strength, we generate images under multiple noise levels. In addition, we also use the VAE of SDv2.1 \cite{rombach2022high}, as it has been widely involved in the training data of existing detectors. To prevent shortcut bias caused by image format differences, all generated images are saved in JPEG format with a quality factor of 95 before evaluation. By restricting both domains to camera-captured images and applying identical generation and evaluation pipelines, we substantially reduce confounding effects from acquisition modality and generation settings, enabling us to assess the impact of domain shift on the model. More details can be found in Appendix \ref{AIGCDoc-Pilot}.

\subsection{Comparison Baselines}
Following the observation in AIGIBench \cite{li2025artificial} that many previous detectors drop to nearly 50\% accuracy under JPEG95 compression, we select recent detectors with relatively stronger robustness for evaluation, including DRCT \cite{chen2024drct}, Effort \cite{yan2024orthogonal}, B-Free \cite{guillaro2025bias}, DDA \cite{chen2025dual}, MIRROR \cite{liu2026mirror}, and OmniAID \cite{guo2025omniaid}. We conduct all experiments using only their official code and released pre-trained models.

\begin{table*}[t]
    \centering
    \small
    \caption{\textbf{Detection Performance Gap between AI-Generated Natural and Document Images on AIGDoc-Pilot.} For each detector, the decision threshold is calibrated to maintain 95\% Real Acc. \graylabel{Gray rows} denote results on document images, while white rows denote results on natural images.}
    \label{tab:detection_gap}
    \resizebox{1.0\linewidth}{!}{
    \begin{tabular}{l|c|c|cccc|cc|cc|cc}
    \toprule
    \multirow{2}{*}{Method}     & \multirow{2}{*}{SDv2.1-VAE}    & \multirow{2}{*}{\makecell{HiDream-\\E1-Full}}  & \multicolumn{4}{c|}{FLUX.1-dev}      & \multicolumn{2}{c|}{Qwen-Image}   & \multicolumn{2}{c|}{SDv3.5-Large} & \multicolumn{2}{c}{Mean}\\ \cmidrule{4-11}\cmidrule{12-13}
                                                                    & ~         & ~     & 0.2   & 0.4   & 0.6   & 0.8   & 0.2   & 0.4   & 0.2   & 0.4       & F.Acc.                        & AUC                       \\ \midrule
    ~                                                               & 13.21     & 15.41 & 6.61  & 9.17  & 17.98 & 29.54 & 11.56 & 17.06 & 10.64 & 12.48     & 14.37                         & \textbf{69.78}            \\
    \rowcolor{light-gray}\cellcolor{white}\multirow{-2}{*}{DRCT}    & 15.23     & 9.54  & 7.89  & 11.56 & 19.82 & 35.96 & 24.06 & 37.19 & 14.63 & 13.70     & $\textbf{18.96}_\text{+4.59}$ & $68.36_\text{-1.42}$      \\ \midrule
    ~                                                               & 1.83      & 1.10  & 0.55  & 2.02  & 4.40  & 7.89  & 4.77  & 4.77  & 1.10  & 0.55      & 2.90                          & \textbf{55.73}            \\
    \rowcolor{light-gray}\cellcolor{white}\multirow{-2}{*}{Effort}  & 1.65      & 1.10  & 1.10  & 2.20  & 5.87  & 15.96 & 10.94 & 12.50 & 0.75  & 0.94      & $\textbf{5.30}_\text{+2.40}$  & $40.46_\text{-15.27}$     \\ \midrule
    ~                                                               & 98.53     & 70.83 & 27.89 & 22.94 & 32.11 & 51.38 & 77.98 & 82.75 & 93.21 & 95.41     & \textbf{65.30}                & \textbf{88.08}            \\
    \rowcolor{light-gray}\cellcolor{white}\multirow{-2}{*}{B-Free}  & 91.38     & 49.36 & 17.06 & 19.63 & 35.60 & 41.83 & 62.81 & 70.62 & 59.85 & 66.79     & $51.49_\text{-13.81}$         & $84.23_\text{-3.85}$      \\ \midrule
    ~                                                               & 96.88     & 67.52 & 40.55 & 49.36 & 62.39 & 63.30 & 84.40 & 84.59 & 88.07 & 89.54     & \textbf{72.66}                & \textbf{92.54}            \\
    \rowcolor{light-gray}\cellcolor{white}\multirow{-2}{*}{DDA}     & 95.23     & 51.01 & 22.94 & 24.04 & 33.76 & 58.17 & 69.06 & 86.25 & 71.11 & 82.55     & $59.41_\text{-13.25}$         & $86.75_\text{-5.80}$      \\ \midrule
    ~                                                               & 43.12     & 44.95 & 13.21 & 56.51 & 75.23 & 90.28 & 54.86 & 65.14 & 44.77 & 73.76     & \textbf{56.18}                & \textbf{81.78}            \\
    \rowcolor{light-gray}\cellcolor{white}\multirow{-2}{*}{MIRROR}  & 50.09     & 40.73 & 12.84 & 24.04 & 44.04 & 55.41 & 50.31 & 93.12 & 39.96 & 83.11     & $49.37_\text{-6.82}$          & $78.05_\text{-3.73}$      \\ \midrule
    ~                                                               & 48.44     & 55.96 & 10.28 & 22.57 & 30.83 & 62.94 & 29.72 & 35.96 & 27.89 & 55.41     & \textbf{38.00}                & \textbf{74.82}            \\
    \rowcolor{light-gray}\cellcolor{white}\multirow{-2}{*}{OmniAID} & 19.08     & 44.40 & 9.36  & 7.71  & 9.72  & 21.28 & 32.50 & 58.13 & 34.52 & 57.79     & $29.45_\text{-8.55}$          & $62.03_\text{-12.79}$     \\
    \bottomrule
    \end{tabular}
    }
\end{table*}

\subsection{Finding 1: A Pronounced Gap between Natural and Document Images}
To ensure a fair comparison across detectors, we calibrate the decision threshold of each model such that its Real Acc. is fixed at 95\%. Under this setting, we report the Fake Acc. and AUC of each detector on AIGDoc-Pilot in Table~\ref{tab:detection_gap}. Ideally, AI-generated image detectors should be largely independent of image semantics, as they are designed to capture generation artifacts rather than category-specific visual content. Under the same generator and generation settings, their detection performance should therefore remain comparable between AI-generated natural images and document images. However, Table~\ref{tab:detection_gap} clearly violates this assumption. Across all evaluated detectors, the mean AUC consistently declines on document images, with an average drop of 7.15\%. The degradation is also evident in F.Acc. for stronger detectors: B-Free and DDA drop by 13.81\% and 13.25\%, respectively. \textbf{These results demonstrate a pronounced domain gap between natural-image and document-image scenarios.} More details of AUC results are shown in Table \ref{tab:detection_gap_auc} of Appendix.

This gap remains observable even for detectors whose training data already involves SDv2.1-related reconstruction or image-conditioned generation. Under the SDv2.1-VAE setting, B-Free, DDA, and OmniAID still show F.Acc. drops of 7.15\%, 1.65\%, and 29.36\%, respectively, on document images, despite the generator being relatively close to their training distribution. In addition, when the noise strength of FLUX.1-dev reaches 0.8, the generated image is almost fully regenerated, and the original semantics are largely weakened. Nevertheless, MIRROR achieves 90.28\% F.Acc. on natural images but only 55.41\% on document images. This suggests that the gap cannot be simply attributed to semantic preservation in image-to-image generation; instead, document images intrinsically affect how generation artifacts are represented and perceived by detectors.

DRCT and Effort appear to be exceptions, as their document-image F.Acc. is slightly higher than that on natural images. However, their detection accuracy is extremely low in both domains, suggesting that they fail to reliably identify generated images. Moreover, their AUC still follows the same trend, decreasing by 1.42\% and 15.27\%, respectively. 

Overall, these results demonstrate that existing detectors do not learn fully domain-independent artifact representations. Even under controlled generation settings, document images consistently introduce additional difficulty, making AI-generated document image detection a distinct and more challenging problem than conventional AI-generated natural image detection.

\subsection{Finding 2: Strong Spatial Inconsistency of Artifacts.}
\label{sec:finding2}

\noindent
\begin{minipage}[t]{0.48\textwidth}
Existing detectors commonly adopt centercrop during image preprocessing, implicitly assuming that the selected crop can provide representative artifact evidence for the entire image. However, the content of document images is often spatially sparse and unevenly distributed. Therefore, we examine whether detectable artifacts in AI-generated document images are uniformly distributed across different local regions.

\vspace{0.7em}

In particular, we investigate both AI-generated natural images and document images. For each image, we extract 25 crop windows arranged on a uniform $5\times5$ grid. The crop size is set to the standard input size of each detector, and the window includes the conventional center crop.

Using AIGDoc-Pilot, we randomly sample 300

\end{minipage}\hfill
\begin{minipage}[t]{0.49\textwidth}
    \vspace{-0.7em}
    \centering
    % \large
    \captionof{table}{\textbf{Spatial Inconsistency of Fake-Score.} We report $\textit{Max}-\textit{Center}$, where \textit{Max} is the largest fake score among sliding-window crops and \textit{Center} is the standard center-crop. All scores are mapped to $[0,1]$ using the Sigmoid function. \graylabel{Gray rows} denote document images, while white rows denote natural images.}
    \label{tab:spatial_delta}
    \resizebox{\linewidth}{!}{
    \renewcommand{\arraystretch}{1.1}
    \begin{tabular}{l|ccccc|c}
    \toprule
    \multirow{2}{*}{Method} & \multirow{2}{*}{\makecell{SDv2.1-\\VAE}} & \multirow{2}{*}{\makecell{HiDream-\\E1-Full}} & \multirow{2}{*}{\makecell{FLUX.1-\\dev}} & \multirow{2}{*}{\makecell{Qwen-\\Image}} & \multirow{2}{*}{\makecell{SDv3.5-\\Large}} & \multirow{2}{*}{Mean} \\
                                                                    &           &           &           &           &           &       \\ \midrule
    ~                                                               & 0.074     & 0.320     & 0.366     & 0.346     & 0.199     & 0.261 \\
    \rowcolor{light-gray}\cellcolor{white}\multirow{-2}{*}{B-Free}  & 0.187     & 0.402     & 0.436     & 0.367     & 0.394     & 0.357 \\ \midrule
    ~                                                               & 0.066     & 0.263     & 0.215     & 0.163     & 0.120     & 0.165 \\
    \rowcolor{light-gray}\cellcolor{white}\multirow{-2}{*}{DDA}     & 0.098     & 0.397     & 0.371     & 0.281     & 0.282     & 0.286 \\ \midrule
    ~                                                               & 0.403     & 0.397     & 0.303     & 0.403     & 0.359     & 0.373 \\
    \rowcolor{light-gray}\cellcolor{white}\multirow{-2}{*}{MIRROR}  & 0.392     & 0.463     & 0.525     & 0.470     & 0.478     & 0.465 \\ \midrule
    ~                                                               & 0.007     & 0.311     & 0.388     & 0.304     & 0.088     & 0.220 \\
    \rowcolor{light-gray}\cellcolor{white}\multirow{-2}{*}{PROBE}   & 0.054     & 0.464     & 0.488     & 0.296     & 0.286     & 0.318 \\ \midrule
    ~                                                               & 0.177     & 0.164     & 0.164     & 0.210     & 0.209     & 0.185 \\
    \rowcolor{light-gray}\cellcolor{white}\multirow{-2}{*}{OmniAID} & 0.201     & 0.258     & 0.117     & 0.210     & 0.266     & 0.211 \\
    \bottomrule
    \end{tabular}
    }
\end{minipage}

\vspace{-0.5em}
images for each generation setting. For each image, we compute the fake scores of all 25 crops and use the difference between the maximum score and the center-crop score to measure spatial inconsistency. As shown in Table~\ref{tab:spatial_delta}, both AI-generated natural and document images exhibit uneven artifact responses, but the effect is more pronounced on document images. On average, the fake-score increases from 0.241 on natural images to 0.327 on document images, yielding an additional gap of 0.086. \textbf{This indicates that fake evidence in AI-generated document images is more localized and less spatially persistent}. In some cases, different regions of the same document image can even lead to inconsistent detector responses. For example, MIRROR shows an average increase of 0.465 on document images, which is large enough to substantially affect threshold-based decisions.

We attribute this inconsistency to the visual composition of document images. Compared with natural images with continuous textures and complex structures, document images are often dominated by homogeneous backgrounds and sparse high-frequency elements. Such sparsity makes artifact evidence highly dependent on local structures, while large background regions provide limited cues and may weaken the overall detector response. This further motivates a more detailed analysis of which local regions contribute to real-fake discrimination.

\subsection{Finding 3: Text Density Enhances Authentic-Synthetic Separability.}
\label{sec:finding3}
To investigate how regional structure contributes to spatial inconsistency, we analyze the effect of text density on detector behavior using AIGDoc-Pilot. Specifically, we randomly sample 300 real document images and all their corresponding generated images. Using a lightweight proxy metric based on text-stroke density, we select four categories of cropped regions: no-, low-, medium-, and high-text-density regions. We then evaluate each detector on these region-level crops under the same 95\% RAcc. protocol. More details about calculating text density are shown in Appendix \ref{appendix_finding3}.

As shown in Table~\ref{tab:text_density_region_real95_facc}, \textbf{text density exhibits a clear positive relationship with real-synthetic separability for most detectors}. Specifically, B-Free, DDA, and MIRROR increase the F.Acc. monotonically from text-free to high-text-density regions, with gains of 15.86\%, 18.50\%, and 8.56\%, respectively. Their AUC values also increase consistently, indicating that text-dense regions improve not only threshold-based accuracy but also the overall ranking ability between real and generated samples. OmniAID is the main exception, showing no consistent improvement with increasing text density. We conjecture that this may be related to its semantic-aware design, which makes its predictions less directly dependent on local text-density variations.

\begin{table*}[t]
    \centering
    \caption{\textbf{Document-Region Fake Acc. under Different Text Densities at 95\% Real Acc.} Each generator reports F.Acc. (\%). We also report the mean F.Acc. and mean AUC across generation settings. Rows under each detector correspond to non-text, low-text, medium-text, and high-text regions.}
    \label{tab:text_density_region_real95_facc}
    \resizebox{1.0\linewidth}{!}{
    \begin{tabular}{l|c|c|c|cccc|cc|cc|cc}
    \toprule
    \multirow{2}{*}{Method} & \multirow{2}{*}{Region} & \multirow{2}{*}{\makecell{SDv2.1-\\VAE}} & \multirow{2}{*}{\makecell{HiDream-\\E1-Full}} & \multicolumn{4}{c|}{FLUX.1-dev} & \multicolumn{2}{c|}{Qwen-Image} & \multicolumn{2}{c|}{SDv3.5-Large} & \multicolumn{2}{c}{Mean} \\
    \cmidrule{5-12}\cmidrule{13-14}
    ~                                                               &      &       &       & 0.2   & 0.4   & 0.6   & 0.8   & 0.2   & 0.4   & 0.2   & 0.4   & F.Acc.                           & AUC \\
    \midrule
                                                                    & No   & 72.33 & 39.00 & 14.67 & 17.00 & 34.00 & 28.00 & 62.67 & 65.67 & 45.67 & 47.67 & 42.67                            & 82.16 \\
                                                                    & Low  & 84.67 & 48.33 & 18.67 & 23.33 & 35.00 & 38.00 & 61.33 & 72.00 & 46.33 & 51.00 & 47.87$_{\text{+5.20}}$           & 83.56$_{\text{+1.40}}$ \\
                                                                    & Mid  & 86.67 & 48.00 & 21.00 & 27.00 & 47.00 & 53.00 & 63.00 & 74.00 & 53.33 & 58.67 & 53.17$_{\text{+10.50}}$          & 85.02$_{\text{+2.86}}$ \\
    \rowcolor{light-gray}\cellcolor{white}\multirow{-4}{*}{B-Free}  & High & 96.33 & 47.00 & 28.00 & 34.00 & 53.00 & 56.67 & 69.67 & 77.00 & 56.33 & 67.33 & \textbf{58.53}$_{\text{+15.86}}$ & \textbf{86.42}$_{\text{+4.26}}$ \\
    \midrule
                                                                    & No   & 80.00 & 47.33 & 14.00 & 16.67 & 13.00 & 28.00 & 62.67 & 66.00 & 51.67 & 66.33 & 44.57                            & 81.67 \\
                                                                    & Low  & 88.67 & 63.33 & 22.00 & 18.33 & 24.33 & 39.67 & 66.33 & 73.33 & 67.33 & 80.67 & 54.40$_{\text{+9.83}}$           & 85.98$_{\text{+4.31}}$ \\
                                                                    & Mid  & 93.00 & 64.33 & 29.33 & 28.00 & 38.33 & 51.67 & 67.67 & 74.33 & 68.00 & 79.67 & 59.43$_{\text{+14.86}}$          & 87.63$_{\text{+5.96}}$ \\
    \rowcolor{light-gray}\cellcolor{white}\multirow{-4}{*}{DDA}     & High & 94.33 & 64.67 & 30.00 & 32.33 & 39.00 & 56.67 & 73.00 & 81.67 & 73.33 & 85.67 & \textbf{63.07}$_{\text{+18.50}}$ & \textbf{88.46}$_{\text{+6.79}}$ \\
    \midrule
                                                                    & No   & 16.33 & 36.33 & 7.00  & 8.67  & 16.33 & 26.00 & 48.00 & 87.00 & 32.00 & 68.00 & 34.57                            & 72.54 \\
                                                                    & Low  & 21.33 & 39.67 & 5.67  & 13.00 & 22.67 & 38.00 & 45.67 & 90.67 & 42.67 & 76.00 & 39.53$_{\text{+4.96}}$           & 73.35$_{\text{+0.81}}$ \\
                                                                    & Mid  & 21.67 & 39.00 & 7.67  & 15.00 & 26.67 & 40.33 & 43.33 & 92.67 & 35.67 & 74.00 & 39.60$_{\text{+5.03}}$           & 73.99$_{\text{+1.45}}$ \\
    \rowcolor{light-gray}\cellcolor{white}\multirow{-4}{*}{MIRROR}  & High & 31.00 & 43.33 & 11.00 & 21.00 & 34.67 & 45.00 & 39.33 & 89.33 & 37.67 & 79.00 & \textbf{43.13}$_{\text{+8.56}}$  & \textbf{74.06}$_{\text{+1.52}}$ \\
    \midrule
                                                                    & No   & 83.00 & 38.33 & 10.33 & 10.00 & 12.00 & 6.00  & 44.33 & 59.00 & 38.00 & 63.00 & 36.40                            & 76.09 \\
                                                                    & Low  & 94.33 & 50.00 & 15.67 & 12.00 & 19.00 & 15.00 & 52.33 & 67.00 & 56.33 & 78.00 & 45.97$_{\text{+9.57}}$           & 79.17$_{\text{+3.08}}$ \\
                                                                    & Mid  & 97.33 & 49.00 & 14.67 & 10.67 & 20.00 & 18.33 & 62.00 & 73.00 & 63.33 & 78.00 & 48.63$_{\text{+12.23}}$          & 80.64$_{\text{+4.55}}$ \\
    \rowcolor{light-gray}\cellcolor{white}\multirow{-4}{*}{PROBE}   & High & 98.67 & 56.67 & 19.33 & 12.67 & 25.67 & 28.33 & 62.67 & 62.67 & 64.00 & 81.67 & \textbf{51.23}$_{\text{+14.83}}$ & \textbf{81.22}$_{\text{+5.14}}$ \\
    \midrule
                                                                    & No   & 20.33 & 42.00 & 7.00  & 9.33  & 9.00  & 15.00 & 27.33 & 42.33 & 28.67 & 56.00 & \textbf{25.70}                   & \textbf{62.80} \\
                                                                    & Low  & 19.00 & 37.33 & 6.00  & 7.00  & 7.00  & 15.00 & 24.00 & 43.33 & 31.00 & 51.67 & 24.13$_{\text{-1.57}}$           & 61.91$_{\text{-0.89}}$ \\
                                                                    & Mid  & 17.67 & 31.67 & 9.67  & 9.67  & 8.00  & 17.67 & 22.00 & 47.00 & 27.33 & 53.00 & 24.37$_{\text{-1.33}}$           & 62.34$_{\text{-0.46}}$ \\
    \rowcolor{light-gray}\cellcolor{white}\multirow{-4}{*}{OmniAID} & High & 21.33 & 34.67 & 9.67  & 8.33  & 9.00  & 16.67 & 24.67 & 44.33 & 28.00 & 52.67 & 24.93$_{\text{-0.77}}$           & 61.66$_{\text{-1.14}}$ \\ 
    \bottomrule
    \end{tabular}
    }
\end{table*}

Interestingly, this observation appears to contrast with TextFake benchmark~\cite{zhang2026textfake}, which reports a 'Text Density Curse', where detection accuracy decreases as text density increases. To better understand this discrepancy, we further report the F.Acc. under the default threshold of 0.5 

\vspace{-0.6em}
\noindent
\begin{minipage}[t]{0.55\textwidth}

in Table~\ref{tab:text_density_region_default05_facc} of the Appendix. Our results show a similar trend to TextFake: denser text regions can lead to lower F.Acc. for several detectors. However, this does not necessarily imply weaker real-synthetic separability. As shown in Table~\ref{tab:text_density_region_real95_facc}, text-dense regions achieve higher F.Acc. and AUC for most detectors. Moreover, the calibrated thresholds generally decrease as text density increases, which is shown in Table~\ref{tab:text_density_thresholds}, indicating a clear threshold shift across text-density levels. Therefore, the apparent degradation under a fixed threshold is largely caused by threshold misalignment rather than the absence of discriminative artifacts. In

\end{minipage}\hfill
\begin{minipage}[t]{0.4\textwidth}
    \vspace{-0.4em}
    \centering
    \captionof{table}{\textbf{Calibrated Thresholds under Different Text-Density Regions.} For each detector, the threshold is selected to maintain 95\% R.Acc. on document regions.}
    \label{tab:text_density_thresholds}
    \resizebox{1.0\linewidth}{!}{
    \renewcommand{\arraystretch}{1.2}
    \begin{tabular}{l|cccc}
    \toprule
    Method      & No        & Low       & Mid       & High      \\ \midrule
    B-Free      & 0.779     & 0.513     & 0.342     & 0.214     \\
    DDA         & 0.483     & 0.384     & 0.344     & 0.309     \\
    MIRROR      & 0.470     & 0.156     & 0.140     & 0.112     \\
    PROBE       & 0.620     & 0.240     & 0.161     & 0.100     \\
    OmniAID     & 0.390     & 0.395     & 0.381     & 0.371     \\
    \bottomrule
    \end{tabular}
    }
\end{minipage}
\vspace{-0.2em}

other words, text density makes fixed-threshold decisions less reliable, but improves the intrinsic separability between real and generated document. More details can be found in Appendix \ref{appendix_finding3}.

In conclusion, higher text density improves the separability between real and AI-generated document images. This further validates our analysis in Sec \ref{sec:finding2}: large homogeneous background regions provide limited forensic cues and weaken detector responses, while discriminative artifacts are more likely to appear around text-related high-frequency structures.

\section{AIGDoc: A Real-World Multi-Generator Benchmark for AI-Generated Document Image Detection}
To enable comprehensive evaluation of AI-generated image detectors in document-centric scenarios, we introduce \textbf{AIGDoc}, a real-world, multi-generator benchmark for AI-generated document image detection. AIGDoc comprises document images collected from diverse practical settings, including camera-captured and born-digital documents, together with AI-generated counterparts produced by multiple advanced generation and editing models. It provides a systematic testbed for evaluating detection performance, cross-generator generalization, and document-specific failure modes.

\subsection{Training Setting}
We select 46k camera-captured document images \footnote{All training data are kept strictly disjoint from the real images used in the test dataset.} collected from real-world scenarios and reconstruct them using the SDv2.1 VAE. The  image pairs are used to train our document-image detector. We adopt PE-Core-L14-336~\cite{bolya2025PerceptionEncoder} as the backbone and fine-tune it with LoRA~\cite{hu2022lora}, setting $r=8$ and $\alpha=16$. During training, we use the Adam optimizer with a batch size of 64. We train three model variants as follows:

\noindent\textbf{PE-LoRA.} We use $336 \times 336$ center-cropped windows during both training and inference. For real images, we use MSCOCO~\cite{lin2014microsoft} and reconstruct them using the SDv2.1 VAE.

\noindent\textbf{PE-LoRA$^{\dagger}$.} This variant keeps the same training data and optimization settings as PE-LoRA, but replaces center cropping with the text-density-based region selection strategy described above. Specifically, we select the region with the highest text density for both training and inference.

\noindent\textbf{PE-LoRA$^{\ddagger}$.} This variant follows the same text-density-based training and inference strategy as PE-LoRA$^{\dagger}$, but is trained using the document training split of AIGDoc.

\subsection{Evaluation Datasets}
To cover diverse image generation settings, AIGDoc is organized into multiple subsets. Each subset contains an equal number of real and synthetic images. Details of the generation models and construction procedures for the evaluation subsets are provided in Appendix~\ref{sec:generation_pipeline_details}.

\noindent\textbf{Real Image Construction.}
We collect document images from a wide range of real-world scenarios \cite{artaud2018find,sun2021spatial,Cheng_2023_CVPR,yu2023icdar,zhang2024exploring,zhang2025dvd,wang2026finixdoc} and manually filter them to retain high-quality camera-captured samples, which cover diverse document types such as receipts, invoices, books, newspapers, and examination papers. The collected documents also span multiple languages, including Chinese, English, and so on, providing broad coverage of practical document appearances. In addition, to enrich the data composition and enable evaluation under substantially different acquisition forms, we also include born-digital document images.

\noindent\textbf{Evaluation Datasets.}
To better approximate real-world forgery scenarios, we consider both text-to-image generation and image editing, which correspond to different data distributions. The visualization is shown in Figure \ref{fig:overall} (b).

\textit{i) Text-to-image Generation.}
To obtain fine-grained and accurate prompts for document synthesis, we first use InternVL3.5-38B \cite{wang2025internvl3_5} to generate detailed descriptions for the original document images. The model is instructed to produce structured outputs containing three aspects: \textit{Overall Description}, \textit{Layout Structure}, and \textit{Image Text}. We further refine these prompts with InternVL3.5-38B to enrich prompt diversity while removing potentially sensitive information. Based on these, we generate document images using a set of recent text-to-image models, including HiDream-O1-Image \cite{cai2026hidream}, Z-Image-Turbo \cite{cai2025z}, GPT-Image-2 \cite{openai2026gptimage2}, Nano-Banana-2 \cite{google2026nanobanana2}, and Seedream-4.5 \cite{bytedance2025seedream45}. Each subset contains 1,000 real images and 1,000 AI-generated images. Representative prompt examples and their corresponding generated images are shown in Fig.~\ref{fig:generate_prompt} in the Appendix.

\textit{ii) Image Editing.}
To simulate realistic document forgery scenarios, we further construct AI-generated document images through image editing. Unlike text-to-image generation, image editing preserves the original document layout, paper texture, camera-captured degradation, and background appearance, while modifying only localized sensitive content. Specifically, we use InternVL3.5-38B \cite{wang2025internvl3_5} to identify candidate text regions and generate editing instructions for critical fields, such as amounts, years, dates, and key terms. Based on these localized prompts, we apply recent image editing models, including FLUX.1-Kontext~\cite{blackforestlabs2025fluxkontext}, FLUX.2-Klein, HunyuanImage-3-Instruct \cite{cao2025hunyuanimage}, Longcat-Image-Edit \cite{team2025longcat}, OmniGen2~\cite{wu2025omnigen2}, Qwen-Image-Edit~\cite{wu2025qwenimage}, Step1X-Edit~\cite{liu2025step1xedit}, Seedream 4.5, Seedream 5.0-Lite \cite{bytedanceseed2026seedream5lite} and Nano-Banan-2. This setting reflects a practical manipulation scenario where forged document images remain visually close to the original real images but contain altered semantic information. Each subset contains 2000 real images and their paired AI-generated counterparts.

\textit{iii) Born-digital image editing.}
Born-digital document images also commonly appear in real-world document workflows. To better cover this practical scenario, we additionally construct edited samples from born-digital documents. Specifically, we apply Qwen-Image-Edit and Nano Banana 2 to modify sensitive textual fields in electronic document images, providing a complementary setting to camera-captured document editing. Each subset contains 1000 real images and their paired AI-generated counterparts.

\begin{table*}[t]
    \centering
    \LARGE
    \caption{\textbf{Detection Accuracy across Different AI-Generated Document Image Settings.} Each detector reports Acc. (\%) for each generation setting.}
    \label{tab:document_generation_benchmark}
    \resizebox{1.0\linewidth}{!}{
    \renewcommand{\arraystretch}{1.2}
    \begin{tabular}{lcccccccccc|gg}
    \toprule
    Generator & \rot{UnivFD} & \rot{DRCT} & \rot{Effort} & \rot{B-Free} & \rot{DDA} & \rot{MIRROR} & \rot{OmniAID} & \rot{PROBE} & \rot{SICA} & \rot{PE-LoRA} & \rot{PE-LoRA$^{\dagger}$} & \rot{PE-LoRA$^{\ddagger}$} \\
    \midrule
    \multicolumn{13}{c}{\textit{Text2Image}} \\
    \midrule
    Z-Image-Turbo     & 49.4 & 61.8 & 85.2 & 52.3 & 61.4 & 49.9 & 49.8 & 55.1 & 44.9 & 68.0 & \textbf{92.8} & \underline{90.5} \\
    HiDream-O1-Image  & 56.5 & 62.1 & 72.7 & \underline{97.0} & 95.2 & 66.8 & 63.2 & \textbf{97.9} & 75.8 & 95.2 & \textbf{97.9} & 93.5 \\
    Seedream 4.5      & 50.1 & 51.7 & 58.2 & 95.2 & 91.8 & 52.8 & 53.3 & \textbf{99.5} & 61.8 & \underline{98.0} & 97.4 & 95.7 \\
    Nano-Banana-2     & 50.0 & 46.2 & 60.2 & \underline{66.0} & 58.9 & 50.4 & 51.0 & 63.9 & 55.4 & 59.8 & \textbf{67.5} & 60.5 \\
    GPT-Image-2       & 50.0 & 49.5 & \underline{74.2} & 51.3 & 47.0 & 49.8 & 50.3 & 51.2 & 47.4 & 49.4 & 59.7 & \textbf{83.9} \\
    \midrule
    \multicolumn{13}{c}{\textit{Image-Edit}} \\
    \midrule
    FLUX.1-Kontext             & 49.8 & 51.2 & 54.5 & 49.6 & 53.7 & 50.0 & 51.2 & 50.7 & 51.9 & 51.5 & \underline{58.9} & \textbf{93.5} \\
    FLUX.2-Klein               & 49.5 & 43.7 & 48.3 & 50.7 & 52.1 & 50.0 & 50.5 & 51.1 & 52.0 & 50.2 & \underline{53.3} & \textbf{83.9} \\
    Hunyuan-3-Instruct         & 49.8 & 57.0 & 61.1 & 72.4 & 95.8 & 54.2 & 66.0 & 96.6 & 64.5 & 66.4 & \underline{97.0} & \textbf{99.5} \\
    LongCat-Image-Edit         & 49.8 & 47.9 & 53.3 & 71.9 & 68.0 & 50.4 & 53.1 & 69.1 & 56.8 & 75.7 & \underline{84.2} & \textbf{99.2} \\
    OmniGen2                   & 56.1 & 57.2 & 58.1 & 81.3 & 65.7 & 72.4 & 66.9 & 73.8 & 68.5 & 72.2 & \underline{83.4} & \textbf{94.1} \\
    Qwen-Image-Edit            & 51.0 & 49.8 & 57.2 & 82.2 & 78.9 & 53.7 & 56.4 & 81.8 & 60.4 & 73.3 & \underline{89.8} & \textbf{97.3} \\
    Step1X-Edit                & 49.6 & 55.5 & 59.0 & 74.3 & 76.2 & 50.6 & 53.7 & 78.5 & 55.5 & 75.1 & \underline{82.3} & \textbf{84.5} \\
    Seedream 4.5               & 51.4 & 62.1 & 54.8 & 97.2 & 94.7 & 50.2 & 53.0 & \underline{99.2} & 53.5 & 97.1 & \underline{99.2} & \textbf{99.5} \\
    Seedream 5.0-Lite          & 52.8 & 62.1 & 48.3 & 95.6 & 85.8 & 50.3 & 52.7 & \textbf{98.9} & 54.4 & 93.6 & 98.0 & \underline{98.1} \\
    Nano-Banana-2              & 49.3 & 45.0 & 54.9 & 49.0 & 50.8 & 50.0 & 54.4 & 50.4 & 51.3 & 50.2 & \underline{57.4} & \textbf{60.6} \\
    \midrule
    \multicolumn{13}{c}{\textit{Digital-Image-Edit}} \\
    \midrule
    Qwen-Image-Edit   & 47.8 & 34.9 & 50.1 & 30.6 & 23.9 & 50.0 & 50.0 & 39.1 & 8.3  & 55.9 & \textbf{75.4} & \underline{65.6} \\
    Nano-Banana-2     & 48.1 & 36.8 & 24.5 & 17.5 & 24.6 & \textbf{51.1} & 50.0 & 36.1 & \underline{50.9} & 49.5 & 50.1 & 50.0 \\
    \midrule
    Average           & 50.6 & 51.4 & 57.3 & 66.7 & 66.1 & 53.1 & 54.4 & 70.2 & 53.7 & 69.5 & \underline{79.1} & \textbf{85.3} \\
    \bottomrule
    \end{tabular}
    }
\end{table*}

\vspace{-0.4em}

\subsection{Evaluation Baselines}
To comprehensively evaluate the detection performance of existing detectors on document images, we selected 9 state-of-the-art detectors, including UnivFD \cite{ojha2023towards}, DRCT \cite{chen2024drct}, Effort \cite{yan2024orthogonal}, B-Free \cite{guillaro2025bias}, DDA \cite{chen2025dual}, MIRROR \cite{liu2026mirror}, OmniAID \cite{guo2025omniaid}, PROBE \cite{cao2026detectors}, and SICA \cite{du2026can}.

\subsection{Detection Performance}
We report the accuracy (Acc.) results on the AIGDoc dataset in  Table~\ref{tab:document_generation_benchmark}. Although existing detectors demonstrate strong generalization on natural-image benchmarks, their performance degrades substantially on AI-generated document images. Most methods achieve an mAcc. below 70\%, indicating that document-centric scenarios remain challenging for general-purpose detectors. Among the evaluated methods, PROBE achieves the highest mAcc. of 70.2\%. Notably, compared with PE-LoRA, incorporating text-density-guided region selection improves mAcc. by 9.6\%, which is consistent with our diagnostic findings: discriminative evidence in AI-generated document images tends to concentrate in text-dense, high-frequency regions, whereas homogeneous background regions provide limited forensic cues. In addition, incorporating document images during training partially narrows the natural-document gap, improving detection accuracy by 6.2\% over PE-LoRA$^{\dagger}$.

Furthermore, SICA, which unifies Deepfake, AI-generated image, image-manipulation, and document-forgery detection within a single model, achieves an mAcc. of only 53.7\% on AIGDoc. Despite being designed to cover document forgery and other forensic domains, it still struggles to detect AI-generated document images. This further confirms a significant gap between AI-generated document detection and the fields of document tampering detection and existing AI-generation detection methods, which also underscores that AI-generated document detection is not simply addressed by combining conventional document-forgery detection with general AI-generated image detection, highlighting its distinct research challenges.

Moreover, born-digital image editing is the most challenging setting, where many detectors exhibit near-chance or severely degraded performance. Specifically, under Qwen-Image-Edit, the performance of DDA and PROBE decreases by 55.0\% and 42.7\%, respectively, falling far below the chance level. These results suggest that detectors trained primarily on camera-captured images may fail to generalize to digital document images, highlighting the need for broader document-centric training and evaluation protocols. The corresponding AUC results are shown in Appendix \ref{sec:appendix_aigcdoc_auc}.

\section{Conclusions, Limitations, and Broader Impacts}

\noindent\textbf{Conclusions.}
Our study shows that AI-generated document image detection remains substantially more challenging than conventional natural-image detection. Under controlled generation settings, AIGDoc-Pilot reveals a pronounced performance gap between natural and document images. Further analysis indicates that generation artifacts in documents are spatially inconsistent and that text-dense regions provide stronger evidence for real-synthetic discrimination. The broader evaluation on AIGDoc confirms that existing detectors remain unreliable across diverse document generation and editing settings. Although text-density-guided region selection and document-based training improve performance, they do not fully resolve this challenge.

\noindent\textbf{Limitations.}
Despite the improvements achieved in this work, reliable detection of AI-generated document images remains unresolved. Our controlled analysis primarily focuses on camera-captured documents, whereas born-digital document editing has proven substantially more challenging for existing detectors. Future work will expand the benchmark to cover a broader range of difficult scenarios, including user interfaces, screenshots, reproduced document images, and other challenging acquisition settings.

\noindent\textbf{Broader Impacts.}
This work reveals the practical limitations of current AI-generated image detectors in document-centric scenarios and identifies the underlying causes of their performance gap relative to natural images. We hope that AIGDoc will foster document-aware evaluation protocols and advance AI-generated image detection beyond conventional natural-image benchmarks toward more reliable real-world use. Nonetheless, a potential ethical concern is that malicious actors may exploit AIGDoc to develop more evasive AI-generated document methods.

\section*{Ethics Statement}
AIGDoc is constructed for research on AI-generated document image detection. We review the licenses of all source datasets and the terms of generation APIs, and release only samples permitted for redistribution and derivative use. We also remove sensitive personal information and prohibit the use of AIGDoc for deceptive document generation.

\section*{AI use statement}
In this work, we used generative AI tools for generating text prompts of image generation and generating AI-generated document images. We have reviewed all AI-assisted work. We take responsibility for the final content of this work, including text, claims or artifacts produced with the aid of generative AI.

\bibliography{iclr2027_conference}
\bibliographystyle{iclr2027_conference}

\clearpage

\appendix
\section{Appendix}

\subsection{More details of AIGDoc-Pilot}
\label{AIGCDoc-Pilot}
The AIGDoc-Pilot dataset contains two subsets: a natural-image subset and a document-image subset. Both subsets consist of camera-captured real images, with 545 images in each category. For each real image, we construct the corresponding synthetic samples using image-to-image generation or VAE reconstruction, while keeping generation settings identical for natural and document images. 

\begin{figure}
    \centering
    \includegraphics[width=1\linewidth]{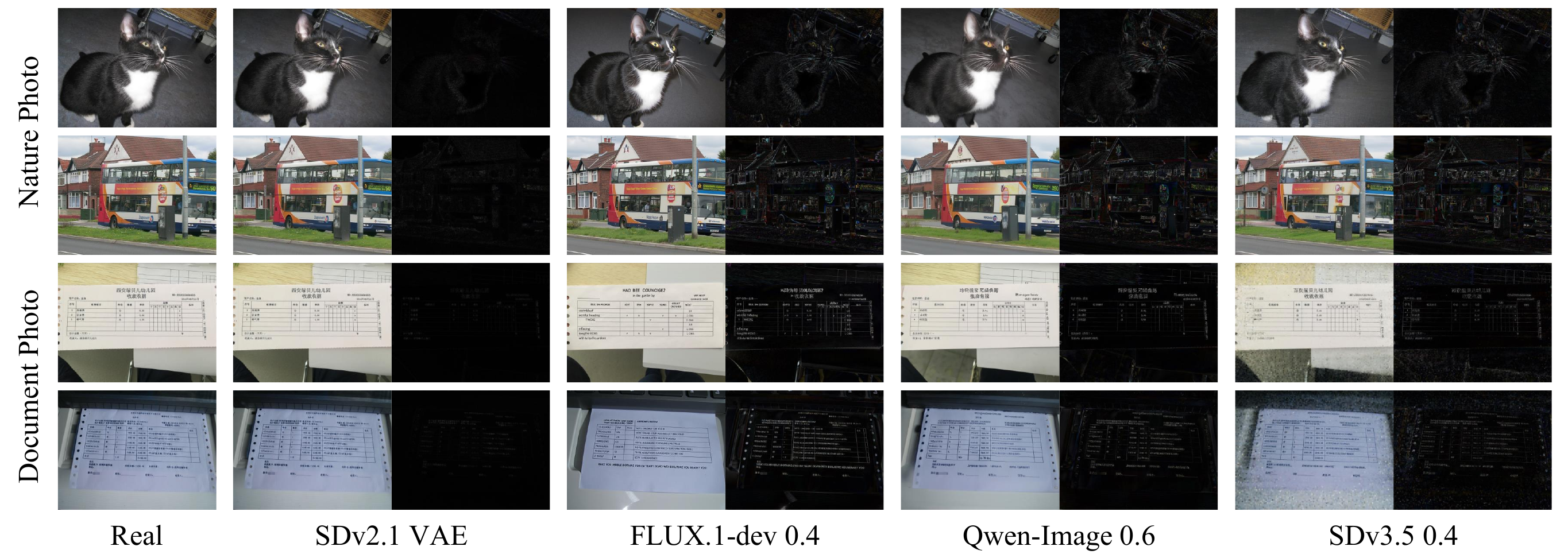}
    \caption{\textbf{Qualitative Visualization of AIGDoc-Pilot Generation Results.} For each real image, we show the generated counterpart and its difference map. Compared with natural images, document images exhibit much sparser generation-induced differences, which are mainly concentrated around text, table lines, and other high-frequency structures.}
    \label{fig:I2I-Diff}
\end{figure}

Figure~\ref{fig:I2I-Diff} provides qualitative examples of the generated images and their pixel-level differences from the corresponding real images. For natural images, the difference maps usually exhibit spatially continuous responses over objects, textures, and backgrounds, indicating that generation-induced changes are distributed across broad visual regions. In contrast, the difference maps of document images are much sparser and mainly concentrate around text characters, table lines, stamps, and other high-frequency structures, while large homogeneous background areas contain only weak changes, highlighting the difficulty of detecting AI-generated document images.

\begin{table*}[t]
    \centering
    \small
    \caption{\textbf{Detection Performance Gap between AI-Generated Natural and Document Images on AIGDoc-Pilot.} Each generator setting reports AUC (\%). \graylabel{Gray rows} denote results on document images, while white rows denote results on natural images.}
    \label{tab:detection_gap_auc}
    \resizebox{1.0\linewidth}{!}{
    \begin{tabular}{l|c|c|cccc|cc|cc|c}
    \toprule
    \multirow{2}{*}{Method}     & \multirow{2}{*}{SDv2.1-VAE}    & \multirow{2}{*}{\makecell{HiDream-\\E1-Full}}  & \multicolumn{4}{c|}{FLUX.1-dev}      & \multicolumn{2}{c|}{Qwen-Image}   & \multicolumn{2}{c|}{SDv3.5-Large} & \multirow{2}{*}{Mean AUC} \\ \cmidrule{4-11}
                                                                    & ~         & ~     & 0.2   & 0.4   & 0.6   & 0.8   & 0.2   & 0.4   & 0.2   & 0.4       &                  \\ \midrule
    ~                                                               & 69.25     & 69.79 & 56.93 & 65.91 & 73.19 & 81.69 & 70.52 & 74.90 & 65.13 & 70.52     & \textbf{69.78} \\
    \rowcolor{light-gray}\cellcolor{white}\multirow{-2}{*}{DRCT}    & 67.88     & 57.76 & 61.23 & 58.91 & 66.63 & 79.09 & 75.30 & 80.94 & 67.28 & 68.54     & $68.36_{\text{-1.42}}$ \\ \midrule
    ~                                                               & 48.80     & 49.02 & 41.53 & 55.24 & 64.54 & 72.91 & 61.74 & 64.30 & 47.95 & 51.31     & \textbf{55.73} \\
    \rowcolor{light-gray}\cellcolor{white}\multirow{-2}{*}{Effort}  & 30.70     & 27.26 & 29.86 & 36.48 & 48.73 & 64.40 & 57.17 & 59.23 & 24.46 & 26.32     & $40.46_{\text{-15.27}}$ \\ \midrule
    ~                                                               & 99.63     & 90.46 & 74.47 & 70.94 & 74.83 & 83.55 & 94.24 & 95.68 & 98.14 & 98.80     & \textbf{88.08} \\
    \rowcolor{light-gray}\cellcolor{white}\multirow{-2}{*}{B-Free}  & 98.01     & 85.08 & 64.55 & 66.13 & 79.96 & 82.81 & 90.83 & 93.66 & 89.08 & 92.17     & $84.23_{\text{-3.85}}$ \\ \midrule
    ~                                                               & 99.31     & 91.12 & 82.24 & 85.71 & 89.70 & 89.71 & 96.22 & 96.31 & 97.49 & 97.62     & \textbf{92.54} \\
    \rowcolor{light-gray}\cellcolor{white}\multirow{-2}{*}{DDA}     & 98.67     & 83.59 & 70.41 & 71.40 & 77.12 & 87.32 & 93.18 & 96.61 & 93.11 & 96.06     & $86.75_{\text{-5.80}}$ \\ \midrule
    ~                                                               & 77.82     & 74.13 & 58.02 & 80.82 & 90.37 & 96.63 & 83.66 & 87.70 & 77.83 & 90.81     & \textbf{81.78} \\
    \rowcolor{light-gray}\cellcolor{white}\multirow{-2}{*}{MIRROR}  & 74.48     & 70.90 & 56.42 & 67.80 & 81.59 & 86.44 & 81.44 & 98.34 & 70.14 & 92.98     & $78.05_{\text{-3.73}}$ \\ \midrule
    % ~                                                               & 99.98     & 96.54 & 85.89 & 78.62 & 81.83 & 87.93 & 97.25 & 98.15 & 99.63 & 99.82     & \textbf{92.56} \\
    % \rowcolor{light-gray}\cellcolor{white}\multirow{-2}{*}{PROBE}   & 99.87     & 94.03 & 79.40 & 73.31 & 78.01 & 78.22 & 96.79 & 97.81 & 96.81 & 99.05     & $89.33_{\text{-3.23}}$ \\ \midrule
    ~                                                               & 84.26     & 82.57 & 54.30 & 62.49 & 68.77 & 85.57 & 72.81 & 77.52 & 73.39 & 86.51     & \textbf{74.82} \\
    \rowcolor{light-gray}\cellcolor{white}\multirow{-2}{*}{OmniAID} & 61.73     & 76.25 & 52.53 & 37.38 & 35.90 & 51.57 & 67.98 & 80.09 & 72.66 & 84.19     & $62.03_{\text{-12.79}}$ \\
    \bottomrule
    \end{tabular}
    }
\end{table*}

\subsection{More Results and Analysis of Finding 1}
\label{appendix_finding1}
We additionally report the AUC of each detector across different generator settings in Table~\ref{tab:detection_gap_auc}. All detectors exhibit lower AUC on document images than on natural images, with an average decrease of 7.15\%, ranging from 1.42\% for DRCT to 15.27\% for Effort. This consistent degradation further reinforces Finding~1: A pronounced performance gap exists between natural and document images.

\begin{figure}
    \centering
    \includegraphics[width=1\linewidth]{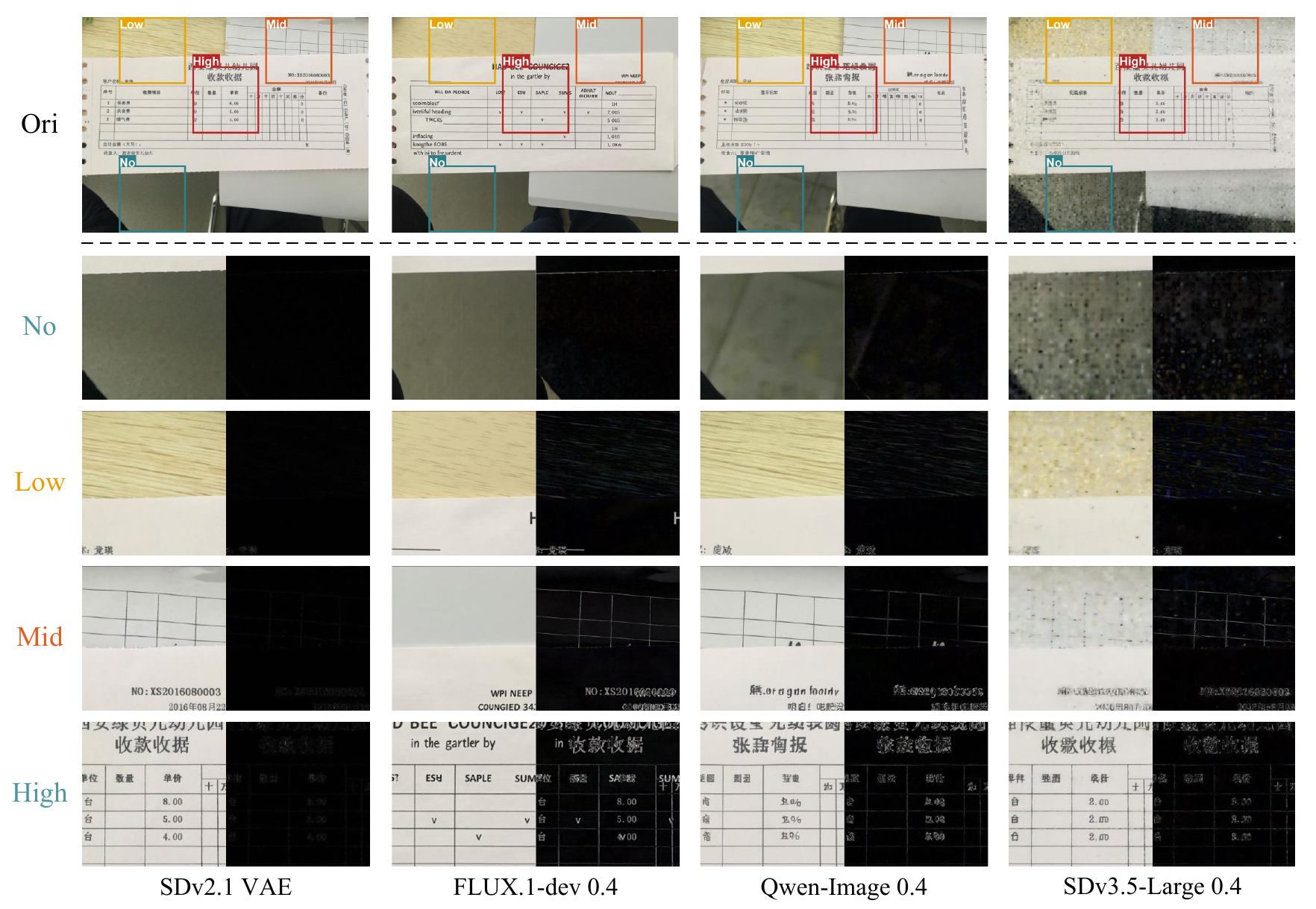}
    \caption{\textbf{Visualization of text-density-based crop regions and their corresponding residual maps.} Low-text-density regions exhibit weaker and less discernible differences between generated and real images, whereas text-dense regions reveal more pronounced generation traces.}
    \label{fig:Density-Diff}
\end{figure}

\subsection{More Results and Analysis of Finding 3}
\label{appendix_finding3}
To examine the effect of text density on detection performance, we develop a region-selection algorithm that identifies four regions with distinct text densities in each document image. Specifically, we sample candidate $448 \times 448$ patches from each real image and compute the proportion of dark, high-gradient pixels within each patch, which serves as a lightweight proxy for dense text strokes and boundaries. Based on this score, we select spatially distinct no-, low-, medium-, and high-text-density regions while limiting their mutual overlap. The selected coordinates are determined solely from the real image and identically applied to its paired generated image, ensuring spatially aligned regional comparisons. Each selected patch is subsequently center-cropped to meet the input requirements of the corresponding detector. As shown in Fig.~\ref{fig:Density-Diff}, the selected regions exhibit clearly distinct text densities. Their residual maps further show that text-dense regions tend to exhibit more pronounced localized differences between real and generated images, suggesting that they may provide more discriminative artifact evidence. 

In addition, we report AUC to assess detector performance independently of a specific decision threshold. As shown in Table~\ref{tab:text_density_region_auc}, the AUC of most detectors consistently increases with text density, indicating that text-dense regions provide stronger discriminative cues for separating real and AI-generated document images. 

However, when using the fixed threshold of 0.5 commonly adopted in existing evaluation protocols, we observe an opposite trend in Table~\ref{tab:text_density_region_default05_facc}: higher text density leads to lower F.Acc., which appears to suggest increased detection difficulty. This discrepancy arises from a score distribution shift across text-density levels. As text density increases, the overall detector scores for both real and generated regions tend to decrease, causing the calibrated threshold for maintaining 95\% Real Acc. to move downward. Therefore, a fixed threshold of 0.5 becomes increasingly misaligned with the actual decision boundary, producing an apparent performance degradation despite improved real-synthetic separability. This is further verified by Table~\ref{tab:text_density_thresholds}, where most detectors require progressively lower thresholds as text density increases.

\begin{table*}[t]
    \centering
    \caption{\textbf{Document-Region AUC under Different Text Densities.} Each generator reports AUC (\%). Rows under each detector correspond to non-text, low-text, medium-text, and high-text regions.}
    \label{tab:text_density_region_auc}
    \resizebox{1.0\linewidth}{!}{
    \renewcommand{\arraystretch}{1.1}
    \begin{tabular}{l|c|c|c|cccc|cc|cc|c}
    \toprule
    \multirow{2}{*}{Method} & \multirow{2}{*}{Region} & \multirow{2}{*}{\makecell{SDv2.1-\\VAE}} & \multirow{2}{*}{\makecell{HiDream-\\E1-Full}} & \multicolumn{4}{c|}{FLUX.1-dev} & \multicolumn{2}{c|}{Qwen-Image} & \multicolumn{2}{c|}{SDv3.5-Large} & \multirow{2}{*}{Mean} \\
    \cmidrule{5-12}
    ~                                                               &      &       &       & 0.2   & 0.4   & 0.6   & 0.8   & 0.2   & 0.4   & 0.2   & 0.4   &  \\
    \midrule
                                                                    & No   & 94.03 & 82.70 & 66.38 & 67.15 & 78.94 & 80.44 & 90.23 & 91.21 & 83.84 & 86.64 & 82.16 \\
                                                                    & Low  & 96.63 & 82.89 & 67.51 & 68.90 & 79.69 & 84.52 & 90.09 & 93.23 & 83.92 & 88.26 & 83.56$_{\text{+1.40}}$ \\
                                                                    & Mid  & 96.76 & 83.63 & 68.63 & 71.98 & 83.36 & 85.43 & 91.09 & 93.32 & 86.24 & 89.73 & 85.02$_{\text{+2.86}}$ \\
    \rowcolor{light-gray}\cellcolor{white}\multirow{-4}{*}{B-Free}  & High & 98.40 & 81.83 & 70.48 & 74.92 & 84.84 & 88.03 & 93.00 & 94.80 & 87.00 & 90.84 & \textbf{86.42}$_{\text{+4.26}}$ \\
    \midrule
                                                                    & No   & 95.78 & 85.87 & 67.37 & 64.15 & 68.58 & 75.90 & 89.94 & 91.51 & 86.22 & 91.34 & 81.67 \\
                                                                    & Low  & 97.88 & 88.59 & 72.88 & 71.67 & 73.89 & 81.94 & 91.90 & 93.97 & 91.88 & 95.20 & 85.98$_{\text{+4.31}}$ \\
                                                                    & Mid  & 98.37 & 89.45 & 76.08 & 74.65 & 76.44 & 86.66 & 92.39 & 94.62 & 92.38 & 95.27 & 87.63$_{\text{+5.96}}$ \\
    \rowcolor{light-gray}\cellcolor{white}\multirow{-4}{*}{DDA}     & High & 98.08 & 88.70 & 74.27 & 76.41 & 80.11 & 87.13 & 93.81 & 95.54 & 94.09 & 96.50 & \textbf{88.46}$_{\text{+6.79}}$ \\
    \midrule
                                                                    & No   & 59.09 & 70.92 & 54.22 & 59.42 & 73.81 & 80.43 & 75.76 & 95.16 & 67.69 & 88.89 & 72.54 \\
                                                                    & Low  & 62.00 & 71.51 & 53.09 & 60.87 & 74.45 & 80.72 & 72.64 & 96.04 & 71.18 & 90.97 & 73.35$_{\text{+0.81}}$ \\
                                                                    & Mid  & 62.37 & 71.89 & 55.51 & 61.03 & 75.66 & 83.26 & 73.56 & 97.59 & 68.72 & 90.29 & 73.99$_{\text{+1.45}}$ \\
    \rowcolor{light-gray}\cellcolor{white}\multirow{-4}{*}{MIRROR}  & High & 64.42 & 72.88 & 55.84 & 61.81 & 75.48 & 82.47 & 72.32 & 95.36 & 67.99 & 92.05 & \textbf{74.06}$_{\text{+1.52}}$ \\
    \midrule
                                                                    & No   & 97.17 & 84.74 & 61.92 & 53.81 & 56.19 & 56.80 & 86.82 & 90.93 & 82.17 & 90.33 & 76.09 \\
                                                                    & Low  & 98.60 & 85.85 & 66.72 & 56.29 & 58.76 & 63.66 & 88.41 & 92.39 & 86.59 & 94.39 & 79.17$_{\text{+3.08}}$ \\
                                                                    & Mid  & 99.22 & 87.57 & 67.89 & 59.74 & 61.81 & 66.88 & 88.43 & 92.42 & 87.98 & 94.43 & 80.64$_{\text{+4.55}}$ \\
    \rowcolor{light-gray}\cellcolor{white}\multirow{-4}{*}{PROBE}   & High & 99.14 & 85.12 & 68.09 & 59.06 & 63.55 & 72.10 & 88.60 & 92.82 & 88.25 & 95.51 & \textbf{81.22}$_{\text{+5.14}}$ \\
    \midrule
                                                                    & No   & 68.31 & 73.02 & 53.70 & 45.51 & 43.18 & 48.44 & 66.62 & 70.71 & 75.10 & 83.44 & \textbf{62.80} \\
                                                                    & Low  & 69.77 & 69.55 & 52.52 & 44.10 & 39.78 & 47.98 & 64.62 & 72.11 & 74.54 & 84.12 & 61.91$_{\text{-0.89}}$ \\
                                                                    & Mid  & 70.47 & 69.74 & 54.08 & 45.19 & 41.69 & 47.84 & 65.13 & 73.59 & 72.72 & 82.93 & 62.34$_{\text{-0.46}}$ \\
    \rowcolor{light-gray}\cellcolor{white}\multirow{-4}{*}{OmniAID} & High & 69.19 & 68.64 & 54.11 & 45.00 & 41.66 & 46.93 & 64.15 & 72.80 & 71.64 & 82.43 & 61.66$_{\text{-1.14}}$ \\
    \bottomrule
    \end{tabular}
    }
\end{table*}

\begin{table*}[t]
    \centering
    \caption{\textbf{Document-Region Fake Acc. under Different Text Densities with the 0.5 Threshold.} Each generator reports F.Acc. (\%). We also report the mean F.Acc. and mean AUC across generation settings. Rows under each detector correspond to non-text, low-text, medium-text, and high-text regions.}
    \label{tab:text_density_region_default05_facc}
    \resizebox{1.0\linewidth}{!}{
    \renewcommand{\arraystretch}{1.1}
    \begin{tabular}{l|c|c|c|cccc|cc|cc|cc}
    \toprule
    \multirow{2}{*}{Method} & \multirow{2}{*}{Region} & \multirow{2}{*}{\makecell{SDv2.1-\\VAE}} & \multirow{2}{*}{\makecell{HiDream-\\E1-Full}} & \multicolumn{4}{c|}{FLUX.1-dev} & \multicolumn{2}{c|}{Qwen-Image} & \multicolumn{2}{c|}{SDv3.5-Large} & \multicolumn{2}{c}{Mean} \\
    \cmidrule{5-12}\cmidrule{13-14}
    ~                                                               &      &       &       & 0.2   & 0.4   & 0.6   & 0.8   & 0.2   & 0.4   & 0.2   & 0.4   & F.Acc.                          & AUC \\
    \midrule
                                                                    & No   & 86.33 & 54.67 & 29.00 & 29.67 & 44.33 & 49.33 & 66.67 & 70.00 & 60.00 & 60.33 & \textbf{55.03}                  & 82.16 \\
                                                                    & Low  & 85.00 & 45.67 & 20.00 & 21.33 & 37.00 & 43.33 & 56.33 & 69.33 & 46.67 & 50.33 & 47.50$_{\text{-7.53}}$          & 83.56$_{\text{+1.40}}$ \\
                                                                    & Mid  & 80.00 & 39.67 & 17.33 & 22.67 & 36.00 & 46.67 & 55.33 & 65.00 & 48.67 & 49.67 & 46.10$_{\text{-8.93}}$          & 85.02$_{\text{+2.86}}$ \\
    \rowcolor{light-gray}\cellcolor{white}\multirow{-4}{*}{B-Free}  & High & 82.00 & 38.00 & 18.00 & 22.33 & 33.00 & 42.00 & 56.00 & 65.33 & 42.33 & 49.33 & 44.83$_{\text{-10.20}}$         & \textbf{86.42}$_{\text{+4.26}}$ \\
    \midrule
                                                                    & No   & 83.00 & 52.67 & 20.00 & 17.33 & 17.33 & 26.00 & 49.33 & 52.00 & 59.00 & 66.00 & 44.27                           & 81.67 \\
                                                                    & Low  & 84.67 & 49.33 & 15.00 & 13.67 & 19.67 & 32.00 & 47.67 & 63.33 & 62.33 & 70.00 & 45.77$_{\text{+1.50}}$          & 85.98$_{\text{+4.31}}$ \\
                                                                    & Mid  & 85.33 & 49.67 & 15.33 & 18.00 & 25.00 & 37.67 & 51.33 & 62.67 & 57.67 & 72.00 & 47.47$_{\text{+3.20}}$          & 87.63$_{\text{+5.96}}$ \\
    \rowcolor{light-gray}\cellcolor{white}\multirow{-4}{*}{DDA}     & High & 88.67 & 48.67 & 18.00 & 16.00 & 23.67 & 38.33 & 57.00 & 67.33 & 62.67 & 74.33 & \textbf{49.47}$_{\text{+5.20}}$ & \textbf{88.46}$_{\text{+6.79}}$ \\
    \midrule
                                                                    & No   & 17.33 & 38.33 & 8.00  & 10.00 & 17.67 & 29.67 & 27.33 & 67.00 & 34.00 & 66.00 & \textbf{31.53}                  & 72.54 \\
                                                                    & Low  & 13.00 & 30.67 & 2.67  & 6.33  & 14.33 & 24.67 & 22.33 & 74.33 & 30.33 & 66.67 & 28.53$_{\text{-3.00}}$          & 73.35$_{\text{+0.81}}$ \\
                                                                    & Mid  & 12.33 & 29.67 & 2.00  & 5.67  & 14.67 & 22.67 & 22.67 & 76.33 & 29.00 & 63.33 & 27.83$_{\text{-3.70}}$          & 73.99$_{\text{+1.45}}$ \\
    \rowcolor{light-gray}\cellcolor{white}\multirow{-4}{*}{MIRROR}  & High & 12.33 & 27.67 & 2.00  & 4.00  & 10.67 & 18.67 & 22.33 & 72.33 & 24.33 & 64.67 & 25.90$_{\text{-5.63}}$          & \textbf{74.06}$_{\text{+1.52}}$ \\
    \midrule
                                                                    & No   & 89.33 & 45.33 & 13.33 & 11.33 & 12.33 & 10.67 & 45.33 & 59.67 & 46.67 & 60.67 & \textbf{39.47}                  & 76.09 \\
                                                                    & Low  & 92.67 & 39.33 & 9.33  & 6.33  & 9.00  & 8.00  & 46.00 & 57.33 & 44.00 & 57.67 & 36.97$_{\text{-2.50}}$          & 79.17$_{\text{+3.08}}$ \\
                                                                    & Mid  & 93.00 & 36.33 & 9.00  & 8.33  & 10.00 & 9.67  & 43.33 & 59.00 & 41.33 & 56.67 & 36.67$_{\text{-2.80}}$          & 80.64$_{\text{+4.55}}$ \\
    \rowcolor{light-gray}\cellcolor{white}\multirow{-4}{*}{PROBE}   & High & 93.00 & 34.33 & 9.33  & 6.00  & 11.67 & 11.33 & 43.33 & 48.67 & 39.33 & 54.67 & 35.17$_{\text{-4.30}}$          & \textbf{81.22}$_{\text{+5.14}}$ \\
    \midrule
                                                                    & No   & 12.67 & 31.33 & 2.67  & 5.00  & 5.00  & 9.33  & 19.33 & 30.33 & 20.00 & 42.00 & \textbf{17.77}                  & \textbf{62.80} \\
                                                                    & Low  & 11.67 & 29.00 & 2.33  & 3.00  & 5.00  & 11.67 & 17.00 & 34.33 & 19.67 & 41.67 & 17.53$_{\text{-0.24}}$          & 61.91$_{\text{-0.89}}$ \\
                                                                    & Mid  & 10.33 & 29.00 & 3.33  & 4.00  & 4.33  & 11.33 & 16.67 & 33.33 & 17.00 & 41.33 & 17.07$_{\text{-0.70}}$          & 62.34$_{\text{-0.46}}$ \\
    \rowcolor{light-gray}\cellcolor{white}\multirow{-4}{*}{OmniAID} & High & 12.33 & 29.67 & 3.33  & 4.00  & 5.33  & 9.00  & 18.00 & 33.67 & 16.33 & 42.67 & 17.43$_{\text{-0.34}}$          & 61.66$_{\text{-1.14}}$ \\
    \bottomrule
    \end{tabular}
    }
\end{table*}

\begin{table*}[!ht]
    \centering
    \LARGE
    \caption{\textbf{Detection AUC across Different AI-Generated Document Image Settings.} Each detector reports AUC (\%) for each generation setting.}
    \label{tab:document_generation_benchmark_auc}
    \resizebox{1.0\linewidth}{!}{
    \renewcommand{\arraystretch}{1.2}
    \begin{tabular}{lcccccccccc|gg}
    \toprule
    Generator & \rot{UnivFD} & \rot{DRCT} & \rot{Effort} & \rot{B-Free} & \rot{DDA} & \rot{MIRROR} & \rot{OmniAID} & \rot{PROBE} & \rot{SICA} & \rot{PE-LoRA} & \rot{PE-LoRA$^{\dagger}$} & \rot{PE-LoRA$^{\ddagger}$} \\
    \midrule
    \multicolumn{13}{c}{\textit{Text2Image}} \\
    \midrule
    Z-Image-Turbo     & 48.0 & 69.5 & 93.2 & 60.9 & 78.4 & 56.4 & 31.8 & 86.1 & 42.9 & 84.2 & \textbf{97.7} & \underline{97.2} \\
    HiDream-O1-Image  & 82.6 & 66.7 & 83.5 & \underline{99.6} & 99.1 & 93.5 & 87.1 & \textbf{99.9} & 78.9 & 99.3 & \textbf{99.9} & 98.7 \\
    Seedream 4.5      & 67.8 & 51.4 & 64.0 & 99.0 & 97.8 & 67.2 & 41.0 & \textbf{100.0} & 64.8 & 99.6 & \underline{99.9} & 99.8 \\
    Nano-Banana-2     & 55.9 & 45.2 & 63.6 & 70.6 & 63.6 & 67.8 & 47.0 & \underline{78.5} & 57.5 & 58.2 & \textbf{79.0} & 64.5 \\
    GPT-Image-2       & 67.7 & 51.0 & \underline{83.3} & 47.3 & 46.9 & 64.4 & 38.2 & 75.8 & 48.3 & 35.8 & 72.7 & \textbf{94.3} \\
    \midrule
    \multicolumn{13}{c}{\textit{Image-Edit}} \\
    \midrule
    FLUX.1-Kontext             & 46.8 & 53.2 & 58.5 & 47.6 & \underline{72.4} & 54.3 & 54.5 & 71.4 & 58.9 & 61.3 & 68.6 & \textbf{99.4} \\
    FLUX.2-Klein               & 43.5 & 42.4 & 36.9 & 44.9 & 54.9 & 50.3 & 54.0 & 51.3 & \underline{58.4} & 41.7 & 41.2 & \textbf{91.8} \\
    Hunyuan-3-Instruct         & 53.3 & 62.1 & 72.6 & 87.3 & 99.4 & 74.8 & 66.8 & \underline{99.7} & 78.7 & 86.6 & \underline{99.7} & \textbf{100.0} \\
    LongCat-Image-Edit         & 44.1 & 49.0 & 55.9 & 87.4 & 87.1 & 61.2 & 58.8 & 89.5 & 70.5 & 94.6 & \underline{96.6} & \textbf{100.0} \\
    OmniGen2                   & 68.8 & 62.9 & 67.2 & 85.4 & 75.2 & 82.7 & 70.6 & 90.3 & 76.9 & \underline{92.1} & 90.3 & \textbf{99.3} \\
    Qwen-Image-Edit            & 55.9 & 51.2 & 64.8 & 92.1 & 94.0 & 68.8 & 72.0 & 96.4 & 70.6 & 89.7 & \underline{97.0} & \textbf{99.8} \\
    Step1X-Edit                & 42.1 & 58.0 & 66.7 & 80.0 & 84.9 & 58.4 & 61.3 & \underline{88.0} & 67.7 & 85.9 & 83.0 & \textbf{97.0} \\
    Seedream 4.5               & 56.3 & 69.1 & 68.5 & 99.4 & 99.4 & 59.5 & 52.4 & \textbf{100.0} & 64.0 & \underline{99.8} & \textbf{100.0} & \textbf{100.0} \\
    Seedream 5.0-Lite          & 62.1 & 68.1 & 58.1 & 98.7 & 97.4 & 57.7 & 52.2 & \textbf{99.9} & 63.7 & 99.2 & \underline{99.8} & \underline{99.8} \\
    Nano-Banana-2              & 32.9 & 44.2 & 67.0 & 48.1 & 63.4 & 60.7 & 57.7 & 74.8 & 56.4 & 56.2 & \textbf{86.6} & \underline{85.5} \\
    \midrule
    \multicolumn{13}{c}{\textit{Digital-Image-Edit}} \\
    \midrule
    Qwen-Image-Edit   & 36.9 & 29.5 & 48.1 & 19.4 & 14.2 & 6.3  & 10.2 & 2.0  & 2.5  & 77.0 & \underline{91.2} & \textbf{99.2} \\
    Nano-Banana-2     & 3.8  & 30.0 & 14.9 & 15.6 & 13.3 & \textbf{61.3} & 54.0 & 0.0  & \underline{57.0} & 11.2 & 28.9 & 13.3 \\
    \midrule
    Average           & 51.1 & 53.1 & 62.8 & 69.6 & 73.0 & 61.5 & 53.5 & 76.7 & 59.9 & 74.8 & \underline{84.2} & \textbf{90.6} \\
    \bottomrule
    \end{tabular}
    }
\end{table*}

\subsection{Details of Compared Detectors}
\label{sec:compared_detectors}

To provide a comprehensive benchmark, we evaluate 9 existing state-of-the-art detection methods, including UnivFD~\cite{ojha2023towards}, DRCT~\cite{chen2024drct}, Effort~\cite{yan2024orthogonal}, B-Free~\cite{guillaro2025bias}, DDA~\cite{chen2025dual}, MIRROR~\cite{liu2026mirror}, OmniAID~\cite{guo2025omniaid}, PROBE~\cite{cao2026detectors}, and SICA~\cite{du2026can}. The compared detectors are summarized as follows.

\noindent\textbf{1) UnivFD} (CVPR 2023)~\cite{ojha2023towards}.
UnivFD investigates the generalization ability of large-scale vision-language representations for fake image detection. It uses CLIP features with simple nearest-neighbor or linear-probing classifiers, showing that pre-trained feature spaces can provide strong cross-generator detection capability without training a detector from scratch.

\begin{table*}[t]
    \centering
    \scriptsize
    \caption{\textbf{Data Scale of AI-Generated Document Images.}}
    \label{tab:document_generation_data_scale}
    \resizebox{1.0\linewidth}{!}{
    \begin{tabular}{l|l|c|c|c|c}
    \toprule
    Evaluation Split    & Method                & Type      & Time      & Resolution        & Data Scale \\
    \midrule
    \multirow{5}{*}{Text2Image}
                        & Z-Image-Turbo         & Open      & 2025.12   & 1024 $\sim$ 1024  & 1,000 \\
                        & HiDream-O1-Image      & Open      & 2026.05   & 2048 $\sim$ 2048  & 1,000 \\
                        & Seedream 4.5          & Closed    & 2025.12   & 1184 $\sim$ 3616  & 1,000 \\
                        & Nano-Banana-2         & Closed    & 2026.02   & 338 $\sim$ 2816   & 1,000 \\
                        & GPT-Image-2           & Closed    & 2026.04   & 724 $\sim$ 2172   & 1,000 \\
    \midrule
    \multirow{10}{*}{Image-Edit}
                        & FLUX.1-Kontext        & Open      & 2025.06   & 512 $\sim$ 1984   & 2,000 \\
                        & FLUX.2-Klein          & Open      & 2026.04   & 336 $\sim$ 1616   & 2,000 \\
                        & Hunyuan-3-Instruct    & Open      & 2026.01   & 527 $\sim$ 1992   & 2,000 \\
                        & LongCat-Image-Edit    & Open      & 2025.12   & 528 $\sim$ 2000   & 2,000 \\
                        & OmniGen2              & Open      & 2025.06   & 336 $\sim$ 3264   & 2,000 \\
                        & Qwen-Image-Edit       & Open      & 2025.08   & 336 $\sim$ 3264   & 2,000 \\
                        & Step1X-Edit           & Open      & 2025.11   & 336 $\sim$ 3264   & 2,000 \\
                        & Seedream 4.5          & Closed    & 2025.12   & 1088 $\sim$ 4096  & 1,992 \\
                        & Seedream 5.0-Lite     & Closed    & 2026.07   & 1088 $\sim$ 4096  & 1,993 \\
                        & Nano-Banana-2         & Closed    & 2026.02   & 271 $\sim$ 1024   & 2,000 \\
    \midrule
    \multirow{2}{*}{Digital-Image-Edit}
                        & Qwen-Image-Edit       & Open      & 2025.08   & 1152 $\sim$ 2464  & 1,000 \\
                        & Nano-Banana-2         & Closed    & 2026.02   & 608 $\sim$ 1024   & 1,000 \\
    \bottomrule
    \end{tabular}
    }
\end{table*}

\noindent\textbf{2) DRCT} (ICML 2024)~\cite{chen2024drct}.
DRCT introduces diffusion reconstruction contrastive training for universal detection of diffusion-generated images. It reconstructs images using diffusion models to generate hard training samples and leverages contrastive learning to encourage the detector to capture diffusion-related artifacts with improved generalization.

\noindent\textbf{3) Effort} (ICML 2025)~\cite{yan2024orthogonal}.
Effort observes that naively fine-tuned detectors can overfit to limited fake patterns, leading to a low-rank and poorly generalizable feature space. It decomposes the feature space into orthogonal subspaces via SVD, preserving the principal pre-trained knowledge while adapting residual components for forgery detection.

\noindent\textbf{4) B-Free} (CVPR 2025)~\cite{guillaro2025bias}.
B-Free proposes a bias-free training paradigm by generating fake images from real images through image-conditioned diffusion models. This construction aligns the semantic content of real and fake samples, encouraging the detector to focus on generation artifacts rather than semantic or dataset biases. It further improves robustness through content-based augmentation.

\noindent\textbf{5) DDA} (NeurIPS 2025)~\cite{chen2025dual}.
DDA demonstrates that pixel-domain alignment alone is insufficient for fully aligning real and synthetic image pairs. Thus, they propose to align synthetic images with real ones across both pixel and frequency domains, thereby mitigating bias.

\noindent\textbf{6) MIRROR} (arXiv 2026)~\cite{liu2026mirror}.
MIRROR reformulates AI-generated image detection as a reference-comparison problem. It projects an input image onto a manifold-consistent real-image reference through sparse linear combination and uses the resulting residuals as detection evidence, aiming to reduce reliance on generator-specific forgery cues.

\noindent\textbf{7) OmniAID} (ICML 2026)~\cite{guo2025omniaid}.
OmniAID aims to decouple semantic content from generation artifacts for universal AIGI detection. It adopts a mixture-of-experts framework with specialized semantic experts and a universal artifact expert, enabling the detector to model content-dependent flaws while preserving content-agnostic artifact representations.

\noindent\textbf{8) PROBE} (ICML 2026)~\cite{cao2026detectors}.
PROBE improves detector generalization by actively exploring difficult regions of the generative space. Instead of treating the generator as a fixed data source, it uses the detector as a critic to guide manifold-level modifications and synthesize hard samples, which are then used to expose failure cases and refine the detector.

\noindent\textbf{9) SICA} (ICML 2026)~\cite{du2026can}.
SICA unifies Deepfake, AI-generated image, image-manipulation, and document-forgery detection within a single detector. To mitigate artifact feature collapse caused by heterogeneous forensic domains, it uses a frozen CLIP backbone as a semantic reference and applies constrained low-rank adaptation to construct a unified yet discriminative artifact feature space.

\subsection{AUC Results on AIGDoc}
\label{sec:appendix_aigcdoc_auc}
To further assess the discriminative ability of detectors on document images, we additionally report AUC as a threshold-independent evaluation metric. As shown in Table~\ref{tab:document_generation_benchmark_auc}, existing detectors exhibit substantial performance variation across different document generation settings. While some methods achieve strong AUC on relatively distinguishable generators, their average performance remains limited, particularly for born-digital image editing.

PROBE achieves the highest average AUC of 76.7\% among the original detectors. Moreover, the text-density-guided variant, PE-LoRA$^{\dagger}$, further improves the average AUC to 84.2\%, yielding a gain of 9.4\% compared with PE-LoRA. This result is consistent with the accuracy analysis in the main paper and further supports our finding that text-dense regions provide stronger discriminative evidence for AI-generated document image detection. In contrast, the low AUC values observed for many methods on Digital-Qwen-Image-Edit and Digital-Nano-Banana-2 confirm that born-digital document editing remains a particularly challenging scenario.

\subsection{Details of Generation Pipeline}
\label{sec:generation_pipeline_details}

Due to the high text density and complex layout structures of document images, simple category-level prompts are insufficient for producing realistic document-like images. We therefore use the large vision-language model InternVL3.5-38B \cite{wang2025internvl3_5} to generate fine-grained prompts from real document images, so that the synthesized images can better preserve document content, layout organization, and realistic forgery scenarios. Due to sensitive-content restrictions, Seedream-4.5 produces only 1,992 edited images.

\noindent\textbf{Statistical Distribution of the Test Set.}
We construct three types of AI-generated document images: text-to-image generation, camera-captured document editing, and born-digital document editing. The generation process involves 13 open-source and closed-source generation settings, producing a total of 27k AI-generated images. Detailed statistics for each generation type and model are shown in Table~\ref{tab:document_generation_data_scale}.

\begin{figure}[t]
    \centering
    \includegraphics[width=1\linewidth]{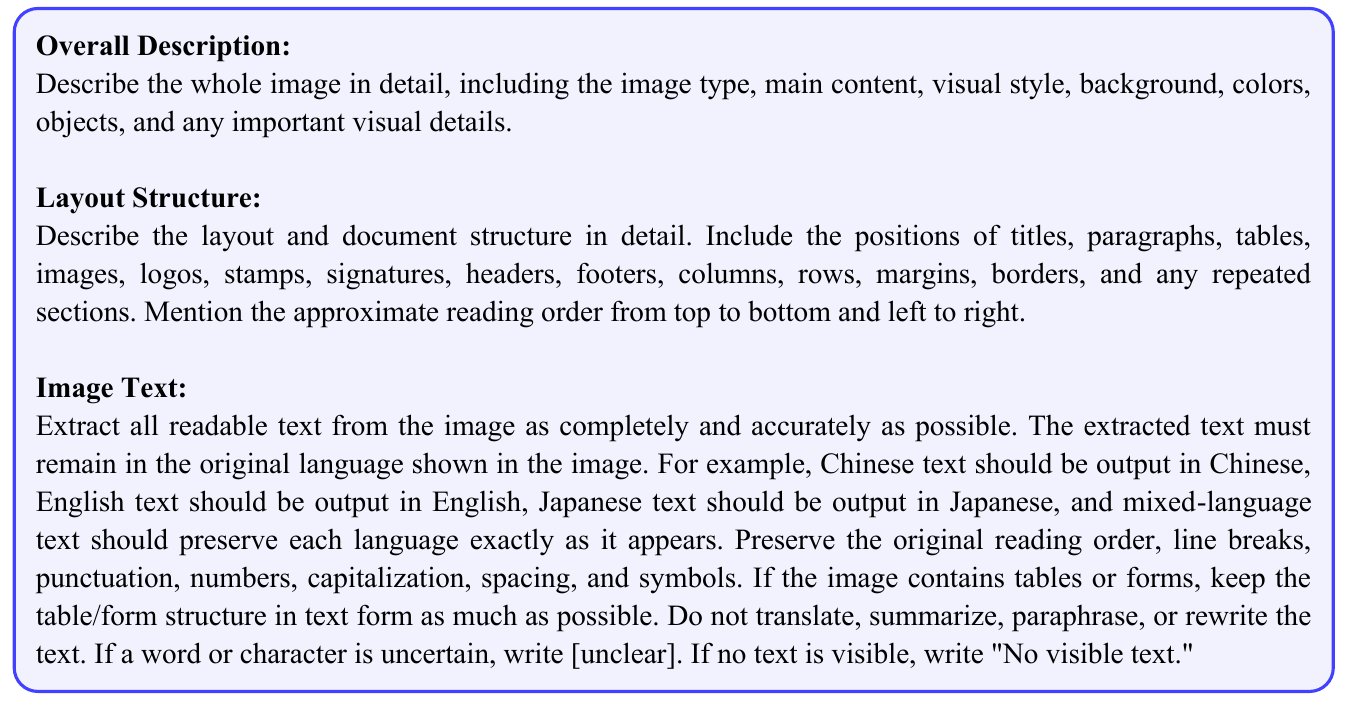}
    \caption{\textbf{Prompt Templates for Image-to-Text Generation.}}
    \label{fig:prompt_generate}
\end{figure}

\begin{figure}
    \centering
    \includegraphics[width=1\linewidth]{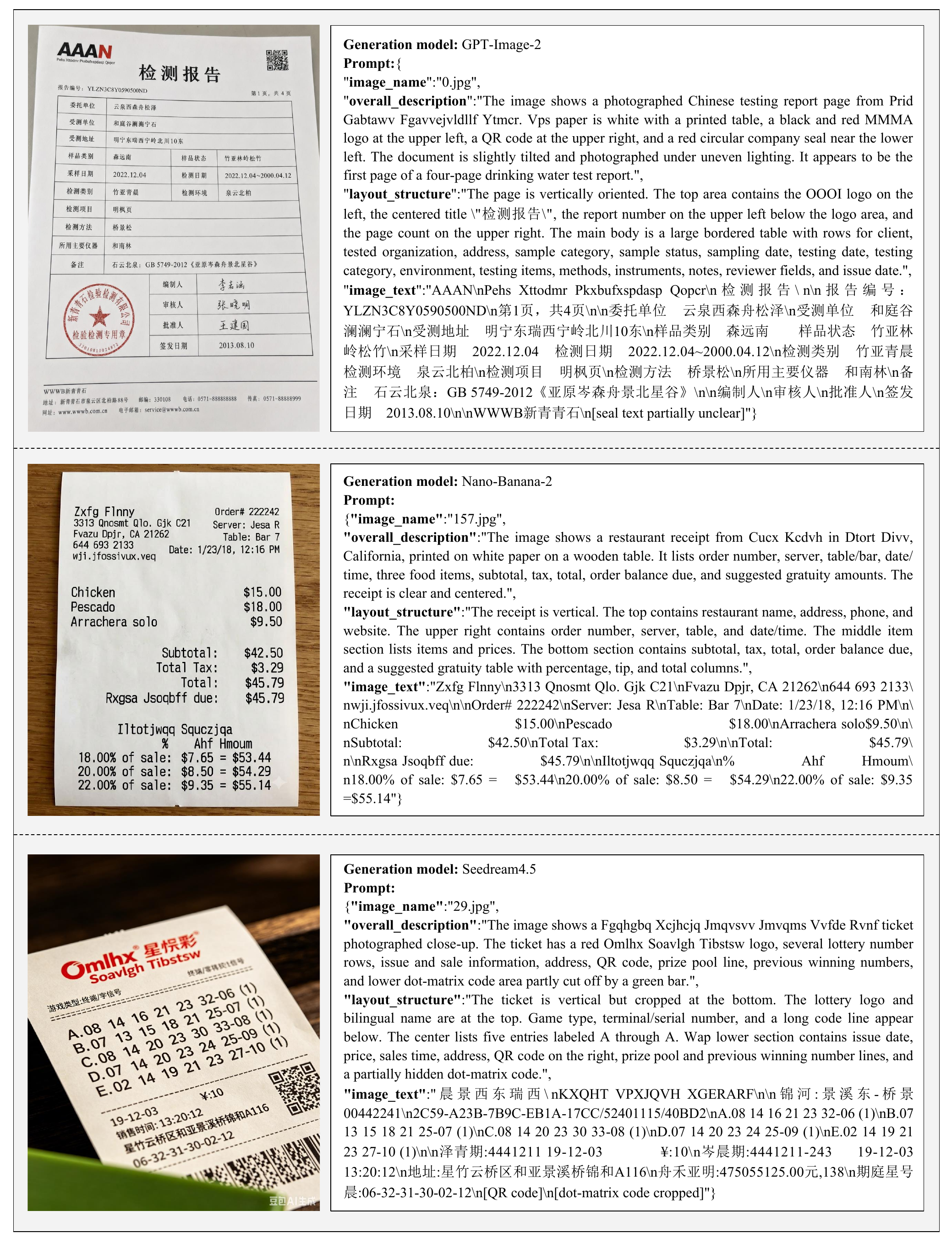}
    \caption{\textbf{Prompt Examples and Generated Images.}}
    \label{fig:generate_prompt}
\end{figure}

\noindent\textbf{Text-to-image Generation.}
We randomly select 600 real document images and use InternVL3.5-38B to generate detailed text prompts for each image. Each prompt follows a three-part structure: \textit{Overall Description}, which characterizes the document type and visual appearance; \textit{Layout Structure}, which specifies the arrangement of text blocks, tables, stamps, and other document elements; and \textit{Image Text}, which records representative visible textual content. To improve prompt diversity while mitigating privacy leakage, we further use InternVL3.5-38B to rewrite and enrich the extracted text, replacing sensitive fields with plausible synthetic information. This process expands the prompt set from 600 to 1,000 entries. The prompt template and representative generated examples are shown in Fig.~\ref{fig:prompt_generate} and Fig.~\ref{fig:generate_prompt}, respectively.

\begin{figure}[t]
    \centering
    \includegraphics[width=1\linewidth]{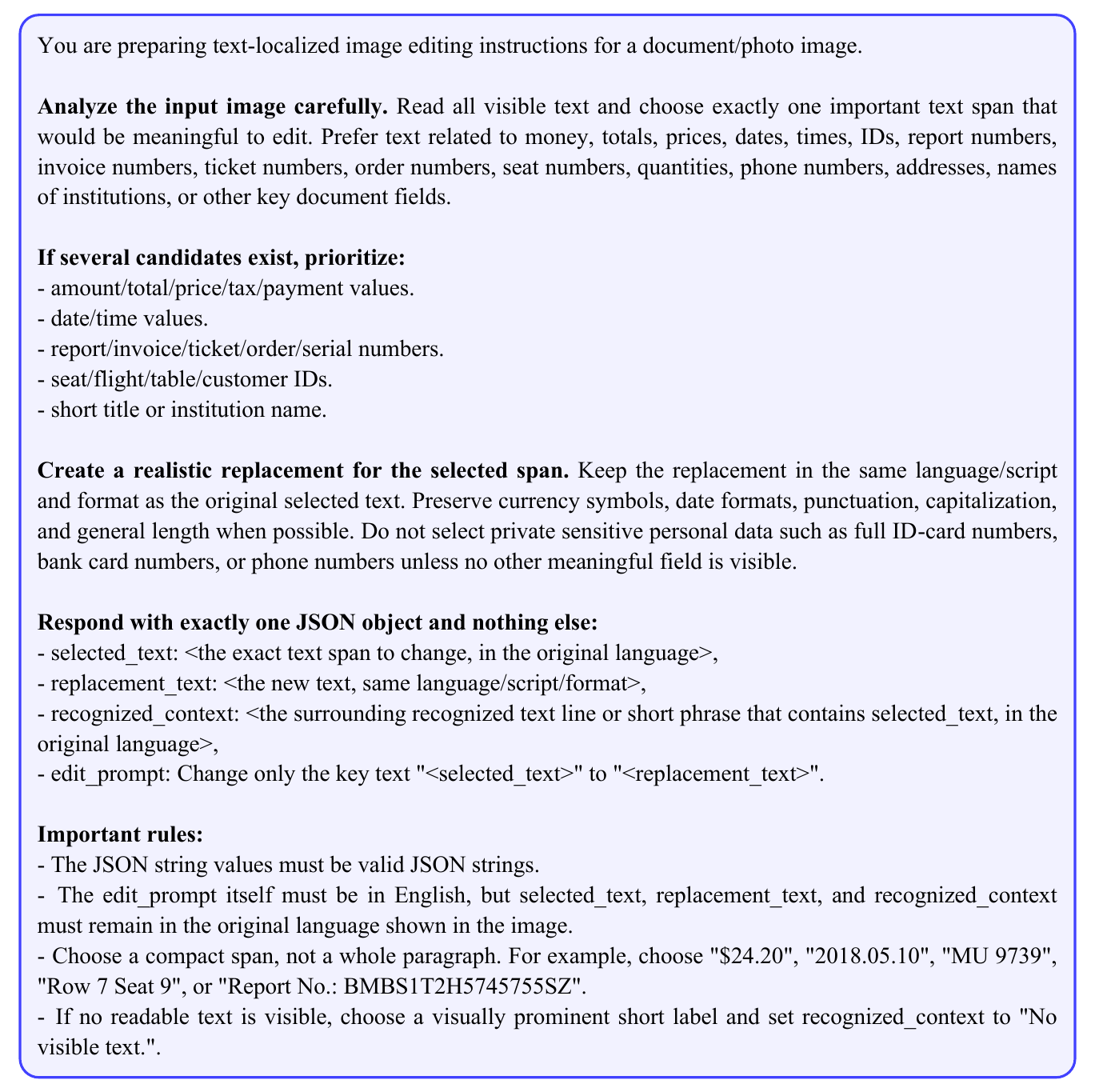}
    \caption{\textbf{Prompt Templates for Image Editing.}}
    \label{fig:prompt_edit}
\end{figure}

\noindent\textbf{Image Editing.}
We use 2K real document images as source images for editing. To better approximate real-world forgery scenarios, we employ InternVL3.5-38B to identify text regions and locate sensitive or editable fields, such as amounts, dates, numbers, and key terms. Based on these recognized fields, we generate localized editing prompts for image editing models. The prompt template used for this process is shown in Fig.~\ref{fig:prompt_edit}.

\end{document}